\documentclass[lettersize,journal]{IEEEtran}
\usepackage{amsmath,amsfonts}
\usepackage{array}
\usepackage[caption=false,font=normalsize,labelfont=sf,textfont=sf]{subfig}
\usepackage{textcomp}
\usepackage{stfloats}
\usepackage{setspace}
\usepackage{url}
\usepackage{verbatim}
\usepackage{graphicx}
\def\BibTeX{{\rm B\kern-.05em{\sc i\kern-.025em b}\kern-.08em
    T\kern-.1667em\lower.7ex\hbox{E}\kern-.125emX}}
\usepackage{balance}

\usepackage[table]{xcolor}
\usepackage{amsmath,amsfonts}
\usepackage{array}
\usepackage[caption=false,font=normalsize,labelfont=sf,textfont=sf]{subfig}
\usepackage{textcomp}
\usepackage{stfloats}
\usepackage{verbatim}
\usepackage{graphicx}
\usepackage{framed}

\usepackage{adjustbox}
\usepackage{pgfplots}
\usetikzlibrary{pgfplots.dateplot}
\usepackage{caption2}
\usepackage{colortbl}
\usepackage{wrapfig}

                                                                   \usepackage{xspace}
\usepackage{threeparttable}
\usepackage{bm}
\usepackage{xcolor}

\usepackage{amsmath}
\usepackage{multirow}
\usepackage{hyperref}
\usepackage[numbers]{natbib}
\usepackage{booktabs}
\usepackage[ruled,vlined]{algorithm2e}
\usepackage{algpseudocode} 

\newcommand{\zqr}[1]{{{\textcolor{red}{[Qianru: #1]}}}}
\newcommand{\qr}[1]{{{\textcolor{red}{[Qianru1: #1]}}}}
\newcommand{\qrr}[1]{{{\textcolor{red}{[Qianru1: #1]}}}}

\newcommand{\zqricml}[1]{{{\textcolor{red}{[Qianru: #1]}}}}
\newcommand{\tkde}[1]{{{\textcolor{red}{[Qianru: #1]}}}}
\newcommand{\proj}{FLDmamba\xspace}

\newcommand{\module}{FMM\xspace}
\renewcommand{\zqr}{}
\renewcommand{\qr}{}
\renewcommand{\qrr}{}
\newcommand{\ICLRRevision}{\color{blue}}
\renewcommand{\ICLRRevision}{}
\renewcommand{\zqricml}{}
\renewcommand{\tkde}{}

\def\model{FLDmamba\xspace}
\def\smodel{FMamba\xspace}

\usepackage[textsize=tiny]{todonotes}

\begin{document}

\title{\emph{\proj:} Integrating Fourier and Laplace Transform Decomposition with Mamba for Enhanced Time Series Prediction}

\author{Qianru Zhang\IEEEauthorrefmark{1}, Chenglei Yu\IEEEauthorrefmark{1}, Haixin Wang, Yudong Yan, Yuansheng Cao, \\Siu-Ming Yiu\IEEEauthorrefmark{5}, Tailin Wu\IEEEauthorrefmark{5}, Hongzhi Yin\IEEEauthorrefmark{5}
\IEEEcompsocitemizethanks{
\IEEEcompsocthanksitem
\IEEEauthorrefmark{1}Equal Contribution,\IEEEauthorrefmark{5}Corresponding author.
\IEEEcompsocthanksitem
Q. Zhang and S.M. Yiu are from the University of Hong Kong. E-mail: \{qrzhang,smyiu\}@cs.hku.hk
\IEEEcompsocthanksitem
C.Yu and T.Wu are from the Westlake University. Email: \{yuchenglei,wutailin\}@westlake.edu.cn
\IEEEcompsocthanksitem
H. Wang is from University of California, Los Angeles. E-mail: whx@cs.ucla.edu.
\IEEEcompsocthanksitem 
Y.Yu and Y.Cao works at Tsinghua University. E-mail: \{yyd21,yscao\}@tsinghua.edu.cn.
\IEEEcompsocthanksitem 
H. Yin works at the University of Queensland. E-mail: h.yin1@uq.edu.au.
}
}

\maketitle

\begin{abstract}\label{sec:abstract}
Time series prediction, a crucial task across various domains, faces significant challenges due to the inherent complexities of time series data, including non-stationarity, multi-scale periodicity, and transient dynamics, particularly when tackling long-term predictions. While Transformer-based architectures have shown promise, their quadratic complexity with sequence length hinders their efficiency for long-term predictions. Recent advancements in State-Space Models, such as Mamba, offer a more efficient alternative for long-term modeling, but they cannot capture multi-scale periodicity and transient dynamics effectively. Meanwhile, they are susceptible to data noise issues in time series. This paper proposes a novel framework, \textbf{FLDmamba} (\textbf{F}ourier and \textbf{L}aplace Transform \textbf{D}ecomposition \textbf{Mamba}), addressing these limitations. \textbf{FLDmamba} leverages the strengths of both Fourier and Laplace transforms to effectively capture both multi-scale periodicity, transient dynamics within time series data, and improve the robustness of the model to the data noise issue. Our extensive experiments demonstrate that \textbf{FLDmamba} achieves superior performance on time series prediction benchmarks, outperforming both Transformer-based and other Mamba-based architectures. To promote the reproducibility of our method, we have made both the code and data accessible via the following URL:{\tkde{ \href{https://github.com/AI4Science-WestlakeU/FLDmamba}{https://github.com/AI4Science-WestlakeU/\model}}}.
\end{abstract}

\begin{IEEEkeywords}
Time Series Data, Mamba, Fourier Transform, Laplace Transform, Time Series Prediction
\end{IEEEkeywords}

\IEEEdisplaynontitleabstractindextext

\IEEEpeerreviewmaketitle

\section{Introduction}
\label{sec:intro}

\IEEEPARstart{T}{ime} series prediction, which forecasts the future values of a (multivariate) variable based on its historical values, finds its application across a wide range of fields. Examples include weather prediction~\citep{lorenc1986analysis,bauer2015quiet}, power grid management~\citep{tang2011smart}, traffic prediction~\citep{yu2017spatio,AGCRN,zhang2025efficient,zhang2024survey,zhang2023automated,zhang2023spatial,zhang2025beyond}, and stock market~\citep{fama1970session,singh2017stock,chen2023tsmixer,bharti2024transformer}, to name just a few. Despite significant advancements in this domain, the inherent complexities of time series data, such as non-stationarity, multi-scale periodicity, intrinsic stochasticity, and noise, pose substantial challenges to existing predictive models in long-term prediction.

\begin{figure}
\vspace*{-4mm}
\centering
\begin{tabular}{cc}
\\\hspace{-4.0mm}
  \begin{minipage}{0.20\textwidth}
	\includegraphics[width=\textwidth]{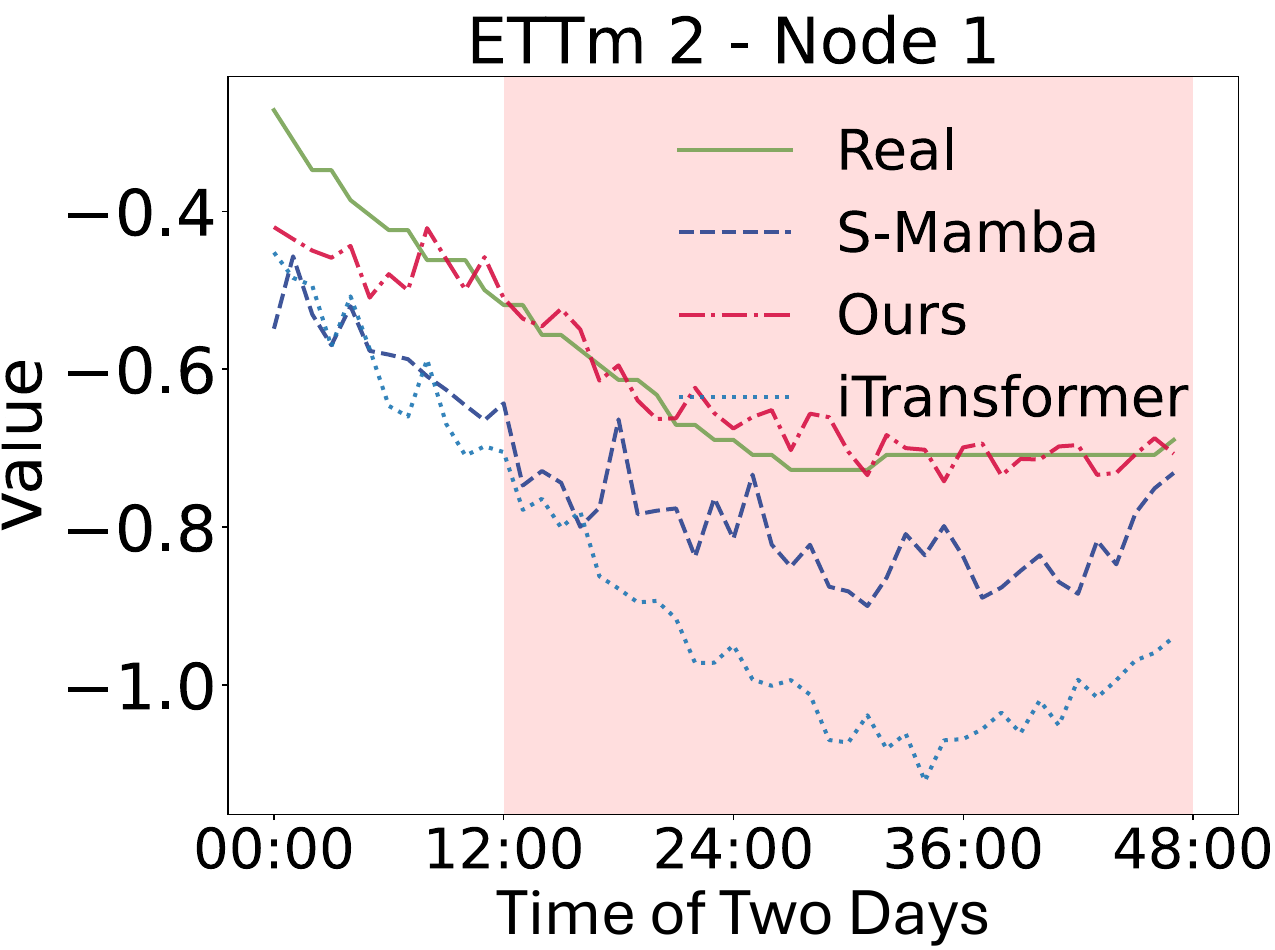}
  \end{minipage}\hspace{-3.mm}
  &
  \begin{minipage}{0.20\textwidth}
	\includegraphics[width=\textwidth]{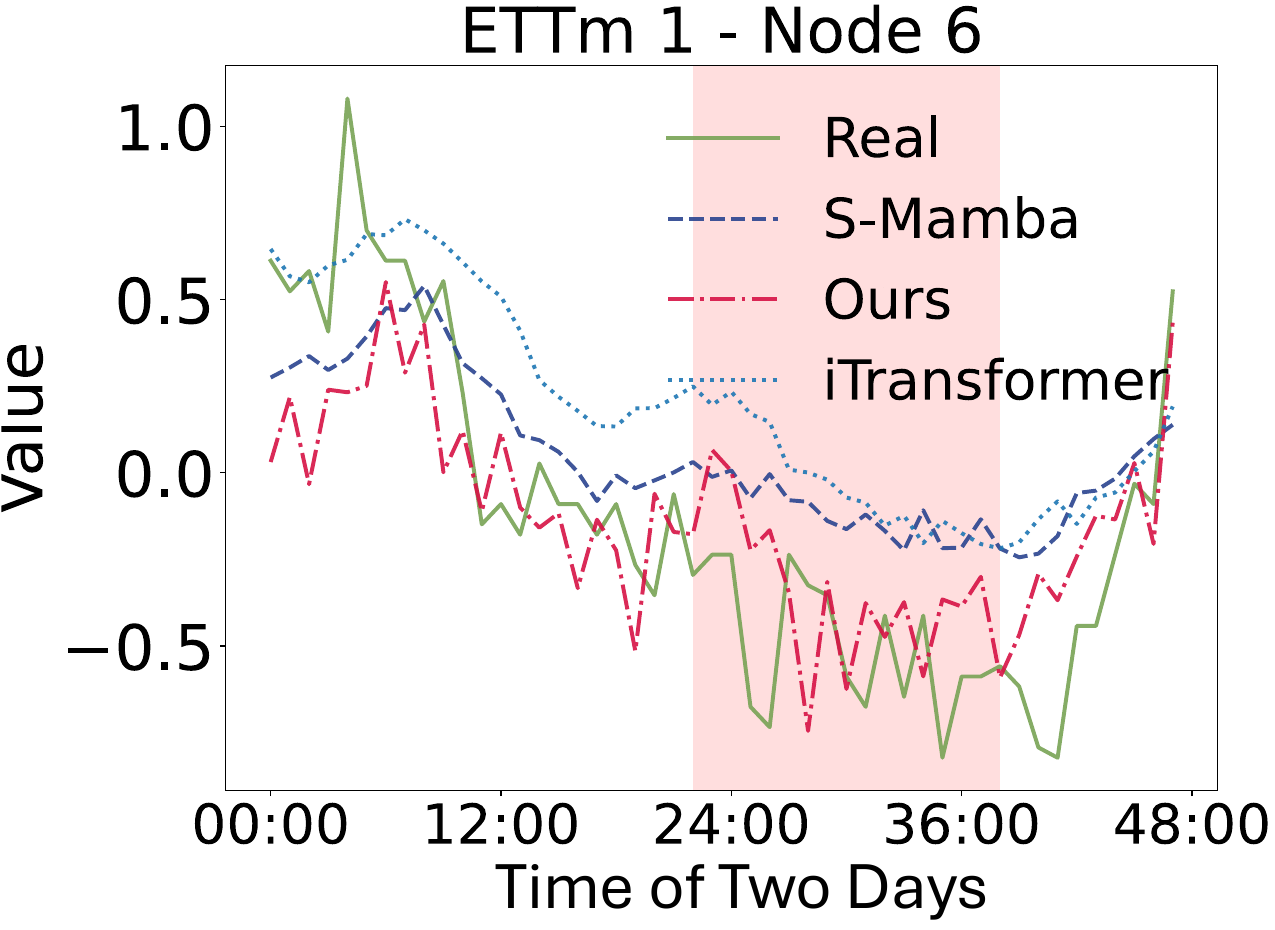}
  \end{minipage}\hspace{-3.0mm}
\end{tabular}
\vspace*{-3mm}
\caption{{Time Series Distributions of ground truth, S-Mamba, iTransformer and Ours.}} %
\vspace*{-2mm}
\label{fig:intro}
\vspace*{-3mm}
\end{figure}

Transformer-based architectures \citep{vaswani2017attention}, successful in NLP and computer vision, have been explored extensively in time series prediction. Although they demonstrate impressive performance, they face degraded accuracy and efficiency in long-term time series prediction due to their quadratic complexity w.r.t. sequence length. iTransformer~\citep{liu2023itransformer} addresses inter-series dependencies by inverting attention layers, but its tokenization approach, which uses a simple MLP layer, fails to capture intricate evolutionary patterns, as shown in Figure~\ref{fig:intro}. Thus, Transformer-based models face challenges in computational efficiency and predictive performance. This can be explained by that the computational cost of self-attention mechanism, which is the core of Transformer-based model, is $\mathcal{O}(L^2)$, where $L$ is the sequence length. Meanwhile, the self-attention mechanism leads to point-wise treatment independently. 

Recently, architectures based on State-Space Models ~\citep{gu2021efficiently,smith2022simplified} have emerged as a promising alternative due to the computational efficiency inherent in linear models to address the long-term prediction challenge. A notable example is Mamba \citep{gu2023mamba}, which employs a linear state space with input-dependent selection. The linear state space allows efficient and parallelized long-sequence modeling, while the input-dependent selection allows propagating and forgetting information in long sequences, facilitating in-context learning. Mamba's design is a good start. However, there are three challenges that Mamba-based methods for time-series prediction cannot address. \textbf{(1) Multi-scale periodicity}. Time series data typically consists of patterns that occur periodically, such as in traffic, electricity, and weather. In addition, the periodic patterns typically exist in multiple time scales and are superimposed together. For example, in weather data, the temperature can fluctuate both in the time scale of a day and a year. Mamba lacks frequency modeling to capture such multi-scale periodicity. \textbf{(2) Transient dynamics}. In addition to periodicity, time series data often shows complex transient dynamics, which can be characterized as time-varying patterns, short-term fluctuations, or event-driven variations. These transient dynamics pose significant challenges for Mamba, as Mamba exhibits a tendency to prioritize point-wise temporal dynamics over neighboring transient dynamics.  Figure~\ref{fig:intro} presents a comparative analysis of the time series predicted by S-Mamba{\tkde{~\citep{wang2025mamba}}} against the ground truth values on the real-world datasets ETTm1 and ETTm2~\citep{haoyietal-informer-2021}. A visual inspection reveals a distinct disparity in the distribution of the predicted time series compared to the ground truth. This discrepancy is due to that S-Mamba fails to effectively capture multi-scale periodicity and transient dynamics inherent within the time series data. \textbf{(3) Data noise.} Noise in time series data introduces random fluctuations into the data, increasing the uncertainty in predictions. Models trained on noisy data may produce less reliable forecasts with wider prediction intervals, making it harder to make accurate predictions.

{\zqr{To address the limitations of existing methods, this paper proposes {\zqricml{a novel framework, \textbf{\model} (\textbf{F}ourier and \textbf{L}aplace Transform \textbf{D}ecomposition \textbf{Mamba}), specifically designed for long-term time series prediction. To be specific, \model\ includes}} two key technical advancements to enhance Mamba in time series prediction: \textbf{(1) Incorporating Fourier analysis into Mamba:} Mamba primarily focuses on capturing temporal dynamics in the temporal domain, lacking the ability to model long-term dynamics in the frequency domain, such as multi-scale patterns overlooked in the temporal domain. To address this, we propose integrating Fourier analysis into Mamba, enabling it to capture long-term properties, such as multi-scale patterns, in the frequency domain. In addition, the Fourier Transform can help in separating the underlying patterns or trends from noise in the time series data by highlighting dominant frequency components. By focusing on these dominant frequencies, the model can reduce the impact of noise that might otherwise affect the accuracy of predictions, thereby enhancing the model's robustness to noisy data. \textbf{(2) Integrating Laplace analysis into Mamba:} To improve Mamba's ability to capture transient dynamics, such as short-term fluctuations, we introduce Laplace analysis into Mamba. This integration allows the model to better understand the relationships between neighboring data points and capture transient changes. {\zqricml{As shown in Figure~\ref{fig:intro}, FLDmamba significantly outperforms S-Mamba. In addition, as the backbone of our framework \model is Mamba, it is highly efficient and well-suited for deployment in large-scale real-world applications.}}

}}

We summarize our contributions as follows:
\begin{itemize}
    \item {\zqr{\textbf{An Efficient Unified Framework for Long-term Time Series Prediction}. We present an efficient and unified framework for long-term time series prediction that eliminates the need for feature engineering.}}
    \item \textbf{Enhanced by Fourier and Laplace Transformations, Decomposed Mamba excels in capturing multi-scale periodicity, transient dynamics, and mitigating noise.} Through the integration of the Fourier and Laplace Transforms into Mamba, our proposed model, \model, adeptly captures intricate multi-scale periodic patterns and dynamic fluctuations present in time series data. This approach not only diminishes the impact of noise but also fortifies the model's resilience, culminating in a substantial enhancement in long-term time series prediction accuracy.
    \item \textbf{Extensive Experiments.} Evaluated on time series prediction benchmarks and comparing with strong baselines including transformer-based and other Mamba-based architectures, our \proj achieves state-of-the-art (SOTA) performance on of tasks. 
\end{itemize}

\vspace{-0.2in}
\section{Related Work}
\label{sec:relate}

\begin{figure*}[htb!]
\centering
\includegraphics[width=0.8\linewidth, height = 0.280\linewidth]{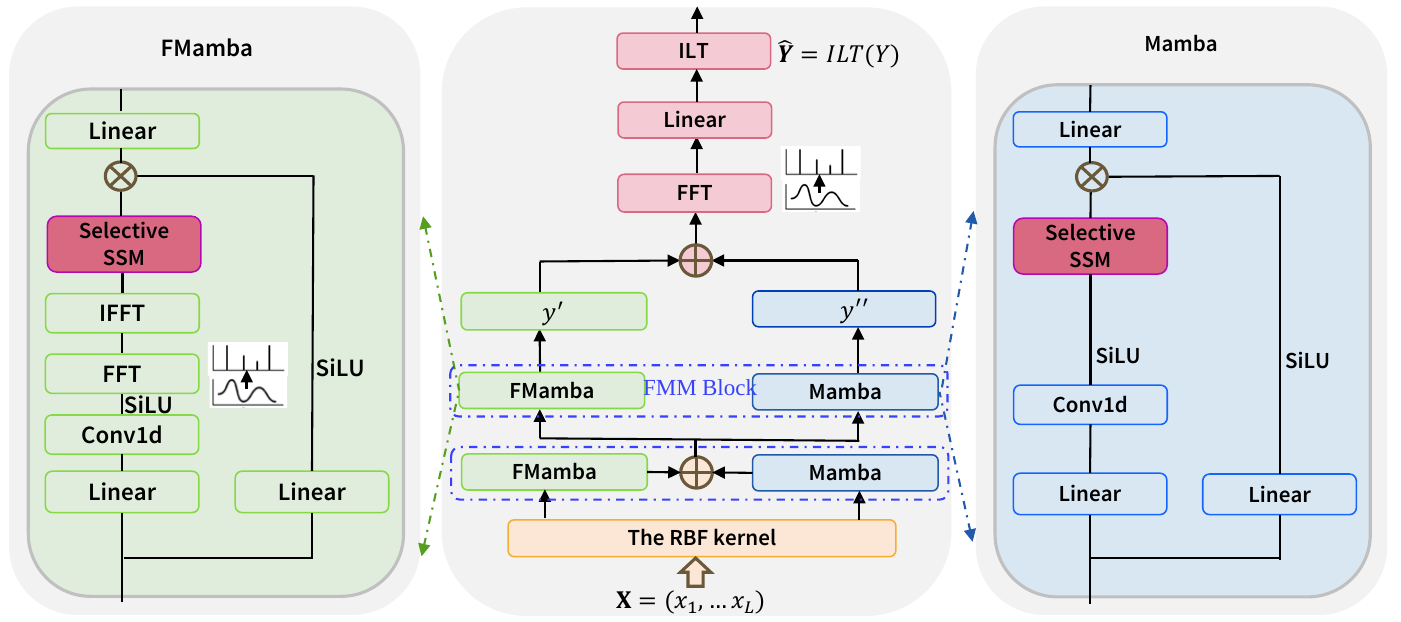}
\vspace{-0.05in}
\caption{{\zqr{This diagram illustrates the architecture of \model, showcasing the individual components and their integration. \textbf{Left}: This section provides a detailed view of the \smodel architecture, highlighting its key components and their interactions. \textbf{Middle:} The central section presents the overall architecture of \model, demonstrating how \smodel, the Fourier and Laplace Transform modules, and Mamba are interconnected to form the complete framework. \textbf{Right:} The rightmost section focuses on the architecture of Mamba, providing a visualization of its internal structure and operation.}}}
\vspace{-0.1in}
\label{fig:fram}
\end{figure*}

\textbf{Time Series Prediction}. Time series prediction, forecasting future values based on historical data~\cite{lim2021time,torres2021deep}, has witnessed a surge in advancements driven by deep neural network techniques. Notably, Mamba~\cite{gu2023mamba} and Transformer~\cite{vaswani2017attention} have emerged as prominent players in this domain, achieving notable successes in time series prediction~\cite{patro2024simba,liang2024bi,vaswani2017attention,zhang2025autohformer}. Transformer-based methods, in particular, have garnered significant attention due to their self-attention mechanism~\cite{vaswani2017attention}, which enables the capture of long-range dependencies within time series data. However, the quadratic complexity inherent in the Transformer architecture presents a formidable challenge for long-term time series prediction. The computational burden associated with processing lengthy sequences significantly hinders the model's performance, particularly when dealing with extended time horizons. This challenge has spurred researchers to explore innovative approaches that balance computational efficiency with predictive accuracy. One such approach, proposed by ~\cite{liu2021pyraformer}, introduces a pyramidal attention module that effectively summarizes features at different resolutions. FEDformer~\cite{zhou2022fedformer} leverages a frequency domain enhanced Transformer architecture to enhance both efficiency and effectiveness.~\cite{zhang2022crossformer} further contribute to this field with Crossformer, which incorporates a patching operation, similar to other models, but also employs Cross-Dimension attention to capture dependencies between different time series. While patching reduces the number of elements to be processed and extracts comprehensive semantics, these models still face limitations in performance when handling exceptionally long sequences. A recent work proposes Moirai \cite{woo2024unified}, which pretrains a model with large-scale datasets. It has different settings from other existing full-shot studies. Thus, it is out-of-scope for our baselines.

To address this persistent challenge, iTransformer~\cite{liu2023itransformer} introduces an innovative approach that inverts the attention layers, enabling the model to effectively capture inter-series dependencies. However, iTransformer's tokenization strategy, which simply passes the entire sequence through a Multilayer Perceptron (MLP) layer, fails to adequately capture the intricate evolutionary patterns inherent in time series data.  This limitation underscores the ongoing need for more sophisticated techniques that can effectively model the complex dynamics of time series data.

In conclusion, while Transformer-based models have demonstrated significant promise in time series prediction, they still grapple with challenges related to computational efficiency and performance when dealing with long sequences.  Continued research efforts are crucial to developing more efficient and effective architectures that can effectively model the intricate complexities of time series data, ultimately paving the way for more accurate and reliable long-term predictions.

\vspace{-0.15in}
\section{Methodology}
\label{sec:solution}

\subsection{Overview and Problem Statement}
\label{sec:overview}
This section details the \model framework (illustrated in Figure~\ref{fig:fram}), which comprises five components: (1) data smoothing using a radial basis function kernel; (2) an FMamba encoder layer (using Fast Fourier Transform for multi-scale periodic pattern extraction); (3) a Mamba encoder layer for modeling long-term dependencies; (4) an integrated FMamba-Mamba block capturing both periodic and transient dynamics and \qrr{separating data noise;} (5) and an inverse Laplace transform to produce time-domain predictions. Then \model's computational complexity is analyzed. Subsequent sections offer a detailed breakdown of each component. Firstly, the problem definition is shown as follows:

\textbf{Problem Statement}.
Given the input with the  long-term time series data $\mathbf{X} = (x_1, ...x_L) \in \mathbb{R}^{L\times V}$, where $L$ is the size of history window and $V$ is number of variates, the ground truth of the predicted output is $\mathbf{Y^{(1)}} = (x_{L+1}, ...x_{L+H})\in \mathbb{R}^{H\times V}$, where $H$ is the prediction size of future time steps. We aim to learn a mapping function $\mathcal{F}$ to satisfy $\widehat{\mathbf{Y}} = \mathcal{F} (\mathbf{X})$ and {\qr{minimize the loss $\frac{1}{|\mathbf{Y^{(1)}}|}\sum_{i=1}^{|\mathbf{Y^{(1)}}|}(\hat{y}_i-y^{(1)}_i)^2$}}, where temporal dependencies are preserved.

\vspace{-0.10in}
\subsection{\model}
\label{sec:model}
Our proposed approach, \model, is illustrated as follows: the Radial Basis Function (RBF) kernel, the \smodel\ encoder layer enhanced by the Fast Fourier transform (FFT), the Mamba encoder layer, the \module block, and the inverse Laplace transform (ILT) module for \model. Each serves a specific purpose in the overall framework. In the following sections, we provide comprehensive explanations and illustrations for each of these components, outlining their respective functionalities and contributions within the \model\ framework. For a comprehensive understanding of the algorithm's steps, please refer to Algorithm~\ref{alg:overall}.


\subsubsection{Data Smoothing via the Radial Basis Function Kernel}
\label{sec:smooth}
To achieve data smoothing on the input data matrix $\mathbf{X}$, we propose the utilization of the radial basis function (RBF) kernel~\cite{scholkopf1997comparing}. The RBF kernel is a widely employed mathematical function in machine learning algorithms, specifically for tasks such as prediction. Its primary purpose is to facilitate the effective capture of intricate temporal relationships and patterns within time series data.

\subsubsection{\smodel\ Encoder Layer Powered by the Fast Fourier Transform (FFT)}

To capture multi-scale periodicity, \emph{e.g.,} daily and monthly patterns, \qrr{and alleviation data noise}, we propose to adopt the Fourier transform to endow state space models on the step size $\Delta \in \mathbb{R}^{2L \times V}$ to filter different periodic patterns out \qrr{from noise}, which is hard to address by existing time-series methods like S-Mamba~\citep{wang2025mamba} and iTransformer~\citep{liu2023itransformer}. In this section, we aim to illustrate the Fourier transform-powered \smodel\ encoder. An important input-dependent selection mechanism is how the step size $\Delta$ is dependent on the input. However, all information of the input is passed through $\Delta$ at each time step without filtering, which has three drawbacks. Firstly, not all information obtained by this selective mechanisms is important. Secondly, after projection via this selective mechanism, the periodic patterns in time series data are hard to capture. \qrr{Thirdly, noise in data is hard to be distinguished by $\Delta$.} Motivated by the above reasons, we propose to adopt the Fourier transform on the $\Delta$ to identify important frequency information and further capture the multi-scale periodic patterns in time series data. Firstly, we define a kernel integral operator, which aims to identify relevant information by convolving the input signal $x$ from the previous layer with a kernel $\tilde{\mathcal{K}}(\Delta t;\phi)$ with time difference $\Delta t$ and  parameter $\phi$:

\textbf{Definition 1: (Kernel integral operator)} We define the kernel integral operator $\mathcal{I}(x;\phi)$ as follows: 
\begin{equation}
\begin{aligned}
    \mathcal{I}(x;\phi)(t)= \int_{D} \tilde{\mathcal{K}}(t-s;\phi) x_{s} ds
\end{aligned}
\end{equation}
Here $t,s$ denote time.
The convolution theorem states that the Fourier transform $\mathcal{F}$ applied to the above kernel integral operator, can be expressed as the product of the Fourier transform of the kernel and the Fourier transform of the input signal. Therefore,
\begin{equation}
\begin{aligned}
    \mathcal{I}(x;\phi)(t)=\mathcal{F}^{-1}(\tilde{W}\cdot \mathcal{F}(x))
\end{aligned}
\end{equation}
Here $\mathcal{F}^{-1}$ is the inverse Fourier transform, $\tilde{W}$ is the Fourier transform of the kernel $\tilde{\mathcal{K}}$, and we directly treat $\tilde{W}$ as a learnable parameter matrix. The functionality of the kernel $\tilde{\mathcal{K}}$ is to identify relevant signals and filter out noise. In addition, to improve the efficiency of operation, Fast Fourier Transform (FFT) is adopted for the above $\mathcal{F}$. For the Fourier transform of the input signal $x$, we define $D=\mathcal{F}(x) \in \mathbb{R}^{2L \times V}$ for each feature $j$ of $x$ as:
\begin{equation}
\begin{aligned}
\label{eq:smamba_fourier}
D_j[k] = \mathcal{F}_j(k) = \sum_{n=1}^{L} x_{nj} \cdot e^{-\hat{i} \frac{2\pi}{L} k n}; j \in [1,2,.. V]; 
\end{aligned}
\end{equation}
$D_j[k] \in \mathbb{C}^{d_f}$ is the Fourier transform of the $j$-th variable at frequency index $k$ and $d_f$ represents the sequence length after FFT in frequency domain. And $\hat{i}=\sqrt{-1}$ denotes the imaginary unit. Then we transform it into temporal domain via Inverse FFT (IFFT),  producing $\Delta_F$, which is the filtered version of $\Delta$, via the kernel integral operator $\mathcal{I}(x;\phi)$ defined above:
\begin{equation}
\begin{aligned}
\label{eq:smamba_dis}
\Delta_F(n,j)&:=\mathcal{I}(x_j;\phi)(n)  = \\
\frac{1}{L} \sum_{k=1}^{L} \tilde{W} &\cdot D_j[k]  \cdot e^{\hat{i} \frac{2\pi}{L} k n};
j \in [1,2,.. V]; \hat{i}=\sqrt{-1}\\
\bar{\mathbf{A}}_F &= \text{exp}(\Delta_{F} \mathbf{A});\\
\bar{\mathbf{B}}_F &= \Delta_{F} \mathbf{A}^{-1}\text{exp}(\Delta_{F} \mathbf{A}) \cdot  \Delta_{F} \mathbf{B}
\end{aligned}
\end{equation}
The filtered $\Delta_F$ replaces the  $\Delta$ in the original Mamba, and can better capture relevant and periodic information in the presence of noise.
Based on the output $\mathbf{X}'$ of the RBF kernel, we can obtain the final output as follows:
\begin{equation}
\begin{aligned}
\label{eq:Fmamba_s2}
u^{(1)}_i &\leftarrow \text{SSM}(\bar{\mathbf{A}}_F, \bar{\mathbf{B}}_F, \mathbf{C})(x'_i);\\
u^{(2)}_i &\leftarrow u^{(1)}_i \otimes \text{SiLU}(\text{Linear}(x'_i));
\end{aligned}
\end{equation}
Where $u_i \leftarrow \text{Linear}(u^{(2)}_i)$ and $x'_i \in \mathbb{R}^{V}$ denotes the output via the RBF at the time step $i$. SiLU denotes the activation function. And Linear represents the linear layer. And $u^{(1)}_i  \in \mathbb{R}^{V}$, $u^{(2)}_i  \in \mathbb{R}^{V}$ and $u_i \in \mathbb{R}^{V}$ are three outputs. A detailed algorithm is shown in Algorithm~\ref{alg:fmodel}.

\subsubsection{Mamba Encoder Layer}{\label{sec:mamba_encoder}} To capture long-term dependencies in time-series sequences, we incorporate Mamba into our framework, working in parallel with \smodel. Unlike the multi-head attention mechanism in Transformer, Mamba employs a selective mechanism to model feature interactions. The core concept of Mamba is to map the input sequence $\mathbf{X}' = (x'_1, x'_2, \ldots, x'_L)$ to the output $U'$ through a hidden state $h(i)$, which acts as a linear time-invariant system. More specifically, given the input sequence $x'_i \in \mathbb{R}^{V}$, where $V$ represents the number of variables in the time series data, we utilize Mamba to model it~\citep{gu2021combining}. The process of Mamba can be outlined as $h'(i)  = \mathbf{A}h(i) + \mathbf{B} x'_i, i \in [1, L]$.
Here, $x'_i \in \mathbb{R}^{V}$. The discretized process, represented by $\Delta$. {\tkde{Here, $\Delta$ represents the time-varying gating mechanism that selectively propagates information based on input-dependent intervals. Intuitively, it functions like a dynamic `timekeeper' that determines when and how much information flows through the system.}} 
\qrr{Then, we can obtain the output via the Mamba encoder layer $U' \in \mathbb{R}^{L \times V}$ as follows:
\begin{equation}
\begin{aligned}
\label{eq:mamba_s2}
u'^{(1)}_i &\leftarrow \text{SSM}(\bar{\mathbf{A}}, \bar{\mathbf{B}}, \mathbf{C})(x'_i);\\
u'^{(2)}_i &\leftarrow u'^{(1)}_i \otimes \text{SiLU}(\text{Linear}(x'_i));
\end{aligned}
\end{equation}
Where $u'_i \leftarrow \text{Linear}(u'^{(2)}_i)$ and $u'^{(1)}_i  \in \mathbb{R}^{V}$, $u'^{(2)}_i  \in \mathbb{R}^{V}$ and $u'_i  \in \mathbb{R}^{V}$ are three outputs at the time step $i$.
}

\subsubsection{The FMamba-Mamba (\module) Block}\label{sec:FMamba} Based on the concepts of \smodel\ and Mamba, we propose the integration of these two components into a single block, which we refer to as the FMamba-Mamba (\module) block. Drawing inspiration from the ResNet mechanism \citep{he2016deep}, an FMM block consists of a FMamba encoder and a Mamba encoder in parallel, both sharing the same input and whose outputs are summed together, producing the output of the FMM block. In this way, it can effectively capture the intricate temporal and periodic dependencies present in the data. Subsequently, the output of the first FMM block is passed to a second \module block (whose output of the second FMamba is $y'$ and output of second Mamba is denoted as $y''$ in Figure~\ref{fig:fram}). Based on this, we have $u''_i = u'_i + u_i$. \qrr{The process is illustrated as follows:
\begin{equation}
\begin{aligned}
\label{eq:fmm_s2}
y'_i &\leftarrow \text{FMamba encoder layer}(u''_i);\\
y''_i &\leftarrow \text{Mamba encoder layer}(u''_i);\\
Y_i &\leftarrow \text{Linear}(\text{FFT}(y'_i+y''_i));
\end{aligned}
\end{equation}
Where $u_i \in \mathbb{R}^{V}$ and $u'_i \in \mathbb{R}^{V}$ denote outputs of the time step $i$ of the first-layer FMamba and the first-layer Mamba respectively. 
$y''_i \in \mathbb{R}^{V}$, $y'_i \in \mathbb{R}^{V}$ and $Y_i \in \mathbb{R}^{V}$.} A detailed description of this process can be found in Figure~\ref{fig:fram} and Algorithm~\ref{alg:overall}. 
Here, FFT preserves the multi-scale nature of solutions through its frequency-domain representation, which is more capable to capture multi-scale temporal patterns based on outputs $y' + y''$ of FMamba and Mamba encoder layers (Fig.~\ref{fig:ablation} and Fig.~\ref{fig:case_study1}). This is followed by Laplace Transform’s exponential kernel $e^{(-st)}$, which would better predict the characteristic oscillations and decays (Fig. ~\ref{fig:ablation} and Fig.~\ref{fig:case_study2}), as will be introduced in the next subsection.
To assess the impact and effectiveness of the \module block, we conducted experiments and present the results in the ablation study.

\subsubsection{Inverse Laplace Transform} \label{sec:ILT}

There are many transient dynamics factors in time series data that hamper the performance of existing methods. Meanwhile, we also aim to capture long-term periodic patterns that are hard to capture in time series data by existing methods. Due to the success of Laplace transform on many domains~\citep{camacho2019laplace}, we propose to adopt the inverse Laplace transform (ILT) on them, which is able to capture transient dynamics and long-term periodic patterns. It is shown as:
{\ICLRRevision{
{\tkde{$\hat{Y}(t) = \frac{1}{2\pi\hat{i}}\lim_{T\to\infty}\int_{\gamma-\hat{i}T}^{\gamma+\hat{i}T} K_\phi(s)Y(s)e^{st}ds; ~~\hat{i} = \sqrt{-1}$}},
where $Y(s)$ is the Laplace transform of $Y(t)$ from the previous layer. And $K_\phi(s)$ is a kernel in the Laplace domain. By stipulating first-order singularities as $K_\phi(s)=\sum_{n=1}^N\frac{\beta_n}{s-\mu_n}$, we derive that
\begin{equation}
\begin{aligned}
\label{eq:llp}
\hat{Y}(t) = \sum_{n=1}^M A_n e^{-\sigma_n t}\cos(w_n t + \varphi_n)
\end{aligned}
\end{equation}
where $A_n, \xi_n, w_n$, and $\phi_n$ are all functions of $Y(t)$ and the $\{\beta_n\}$ and $\{\mu_n\}$. Thus in our work, we directly parameterize $A_n, \xi_n, w_n$, and $\phi_n$ as learnable functions of $Y(t)$ from the previous layer to improve efficiency and stability.
}}We see in Eq.~\ref{eq:llp}, the cosine term $\cos(w_n t)$ plays a crucial role in capturing the periodicity inherent in the data. It is capable of effectively identifying and modeling recurring patterns or cycles within the time series. On the other hand, the term $e^{\sigma_n t }$ is responsible for capturing the transient dynamics exhibited by the data. It enables the model to capture and represent the short-lived variations or irregularities in the time series.
Besides, the combined use of exponential $e^{\sigma_n t }$ and $\cos(w_n t)$ terms ensures that the reconstructed time-domain data accurately reflects both transient dynamics and long-term periodic trends, making it suitable for forecasting future behaviors based on historical data.
This, in turn, contributes to improved accuracy and predictive capabilities, allowing the model to make more reliable forecasts and capture the nuances of the data more effectively.

\subsection{Model Complexity}
\label{sec:complex}
This section presents a complexity analysis of our proposed model, \model. The computational complexity of the base Mamba model is $\mathcal{O}(BLVN)$, where $B$ represents the batch size, $L$ denotes the sequence length, $V$ signifies the number of variables, and $N$ indicates the state expansion factor. The Fast Fourier Transform (FFT) in FLDmamba has time complexity of $\mathcal{O}(BLN\log{L})$, and the inverse Laplace transform has time complexity of $\mathcal{O}(BLN)$, both significantly smaller than $\mathcal{O}(BLVN)$. Therefore, the total time complexity is still $\mathcal{O}(BLVN)$. In other words, \model maintains a comparable computational time complexity to the base Mamba model, making it a promising framework for large-scale real-world applications in time series prediction. This computational efficiency allows \model to handle extensive datasets and complex time series scenarios without significant performance degradation.

\begin{algorithm}[h]
    \caption{The \textbf{\model} Algorithm}
    \label{alg:overall}
    \KwIn{
        \textbf{X}: (B, L, V);
    }
    \KwOut{$\widehat{\mathbf{Y}}$: (B, L, V);\\
    }
    $ U \gets$ \textbf{\smodel}(\textbf{X});  \ \ \ \ // Step into \smodel\ algorithm~\ref{alg:fmodel}\\
    $U' \gets$ \textbf{Mamba}(\textbf{X});\ \ \ \ \ \ // Step into the Mamba algorithm\\
    $U'' \gets U' + U$;\\
    $ y' \gets$ \textbf{\smodel}($U''$);\ \ \ // Step into \smodel\ algorithm~\ref{alg:fmodel};\\
    $ y'' \gets$ \textbf{Mamba}($U''$);\ \ \ \ // Step into Mamba algorithm;\\
    $Y \gets \text{FFT}(y' + y'')$;\\
    $Y \gets \text{Linear}(Y) $;\\
    $\widehat{\mathbf{Y}} \gets \text{ILT}(Y)$;\ \ \ \ // Inverse Laplace Transform module\\
    \textbf{return} $\widehat{\mathbf{Y}}$; 
\end{algorithm}

\begin{algorithm}[h]
    \caption{The \textbf{\smodel} Algorithm}
    \label{alg:fmodel}
    \KwIn{
        \textbf{X}: (B, L, V);
    }
    \KwOut{$U$:(B, L, V);\\
    }
    $\mathbf{X}' \gets$  RBF($\mathbf{X}$);\\
    \For{$p = 1, 2,..., \smodel\ \text{layers}$}{
    \textbf{A}: (V, N) $\leftarrow$ Parameter\\
    \textbf{B}: (V, L, N) $\leftarrow s_{B}(\mathbf{X}')$\\
    \textbf{C}: (B, L, N) $\leftarrow s_{C}(\mathbf{X}')$\\
    $\Delta$: (B, L, N) $\leftarrow$ $\tau_{\Delta}$(Parameter +  $s_{\Delta}(\mathbf{X}')$)\\
    $\Delta' = \text{FFT}(\Delta)$\\
    $\Delta_F = \text{IFFT}({\ICLRRevision{\tilde{W}\cdot}}\Delta')$\\
    $\bar{\mathbf{A}}_F, \bar{\mathbf{B}}_F$ : (B, L, V, N) $\leftarrow discretize(\Delta_F, \textbf{A}, \textbf{B})$\\
    $U^{(1)}$ $\leftarrow$ SSM ($\bar{\mathbf{A}}_F, \bar{\mathbf{B}}_F, \mathbf{C}$)($\mathbf{X}'$)\\
    $U^{(2)}$ $\leftarrow$ $U^{(1)} \otimes \text{SiLU}(Linear(\mathbf{X}'))$\\
    $U$ $\leftarrow$ $Linear(U^{(2)})$
    }
    \textbf{return} $U$; 
\end{algorithm}

\begin{table*}[htb!]
\vspace{-0.1in}
  \caption{We present comprehensive results of \model\ and baselines on the ETTh1, ETTh2, Electricity, Exchange, Weather, and Solar-Energy datasets. The lookback length $L$ is fixed at 96, and the forecast length $T$ varies across 96, 192, 336, and 720. Bold font denotes the best model and underline denotes the second best. All baseline results are obtained from \cite{wang2025mamba} and \cite{woo2024unified}.}
  \label{tab:results_other}
  \renewcommand{\arraystretch}{1.0}
  \centering
  \resizebox{\textwidth}{!}{
  \begin{small}
  \setlength{\tabcolsep}{2.6pt}
  \vspace{1mm}
  \begin{tabular}{c|c|cc|cc|cc|cc|cc|cc|cc|cc|cc|cc|cc|cc|cc}
    \toprule
    \multicolumn{2}{c|}{Models} & \multicolumn{2}{c|}{\textbf{\model (Ours)}}& \multicolumn{2}{c|}{\textbf{S-Mamba}} &  \multicolumn{2}{c}{{\ICLRRevision{SST}}}&  \multicolumn{2}{c}{{\ICLRRevision{Bi-Mamba+}}}& \multicolumn{2}{c}{iTransformer} & \multicolumn{2}{c}{RLinear} & \multicolumn{2}{c}{PatchTST} & \multicolumn{2}{c}{Crossformer} & \multicolumn{2}{c}{TiDE} & \multicolumn{2}{c}{TimesNet} & \multicolumn{2}{c}{DLinear} & \multicolumn{2}{c}{FEDformer} &  \multicolumn{2}{c}{Autoformer} \\
    \cmidrule(lr){1-2}\cmidrule(lr){3-4}\cmidrule(lr){5-6}\cmidrule(lr){7-8} \cmidrule(lr){9-10}\cmidrule(lr){11-12}\cmidrule(lr){13-14}\cmidrule(lr){15-16}\cmidrule(lr){17-18}\cmidrule(lr){19-20}\cmidrule(lr){21-22}\cmidrule(lr){23-24}\cmidrule(lr){25-26}\cmidrule(lr){27-28}   
    \multicolumn{2}{c|}{Metric} & MSE & MAE &  MSE & MAE & \ICLRRevision{MSE} & \ICLRRevision{MAE} & \ICLRRevision{MSE} & \ICLRRevision{MAE}& MSE & MAE & MSE & MAE & MSE & MAE & MSE & MAE & MSE & MAE & MSE & MAE & MSE & MAE & MSE & MAE & MSE & MAE\\
    \toprule
     \multirow{5}{*}{\rotatebox{90}{ETTm1}} 
    &  96 &\textbf{0.318} &\textbf{0.360}&{0.333} & {0.368} &\ICLRRevision{0.337}&\ICLRRevision{0.374}&\ICLRRevision{0.355}&\ICLRRevision{0.386}& 0.334 & 0.368 & 0.355 & 0.376 & \underline{0.329} & \underline{0.367} & 0.404 & 0.426 & 0.364 & 0.387 & 0.338 & 0.375 & 0.345 & 0.372 & 0.379 & 0.419 & 0.505 & 0.475\\
    
    & 192 &\textbf{0.365}&\textbf{0.384}& 0.376 & 0.390 &\ICLRRevision{0.377}&\ICLRRevision{0.392}&\ICLRRevision{0.415}&\ICLRRevision{0.419}& 0.377 & 0.391 & 0.391 & 0.392 & \underline{0.367} & \underline{0.385} & 0.450 & 0.451 & 0.398 & 0.404 & {0.374} & {0.387} & 0.380 & 0.389 & 0.426 & 0.441 & 0.553 & 0.496\\
    
    & 336 &\ICLRRevision{0.404}&\textbf{0.409}& 0.408 & 0.413&\ICLRRevision{\underline{0.401}}&\ICLRRevision{0.412}&\ICLRRevision{0.450}&\ICLRRevision{0.442} & 0.426 & 0.420 & 0.424 & 0.415 & \textbf{0.399} & \underline{0.410} & 0.532 & 0.515 & 0.428 & 0.425 & {0.410} & {0.411} & 0.413 & 0.413 & 0.445 & 0.459 & 0.621 & 0.537\\
    
    & 720 &\underline{0.464}&\underline{0.441}& 0.475 & {0.448} &\ICLRRevision{0.498}&\ICLRRevision{0.464}&\ICLRRevision{0.497}&\ICLRRevision{0.476}& 0.491 & 0.459 & 0.487 & 0.450 & \textbf{0.454} & \textbf{0.439} & 0.666 & 0.589 & 0.487 & 0.461 & 0.478 & 0.450 & {0.474} & 0.453 & 0.543 & 0.490 & 0.671 & 0.561\\
    \cmidrule(lr){2-28}

    & Avg &\underline{0.389}&\textbf{0.399}& {0.398} & {0.405} &\ICLRRevision{0.413}&\ICLRRevision{0.411}&\ICLRRevision{0.429}&\ICLRRevision{0.431}& 0.407 & 0.410 & 0.414 & 0.407 & \textbf{0.387} & \underline{0.400} & 0.513 & 0.496 & 0.419 & 0.419 & 0.400 & 0.406 & 0.403 & 0.407 & 0.448 & 0.452 & 0.588 & 0.517\\
    \midrule
    
    \multirow{5}{*}{\rotatebox{90}{ETTm2}} 
    & 96  &\textbf{0.173}&\textbf{0.253}& {0.179} & {0.263}&\ICLRRevision{0.185}&\ICLRRevision{0.274}&\ICLRRevision{0.186}& \ICLRRevision{0.278}  & 0.180 & 0.264 & 0.182 & 0.265 & \underline{0.175} & \underline{0.259} & 0.287 & 0.366 & 0.207 & 0.305 & 0.187 & 0.267 & 0.193 & 0.292 & 0.203 & 0.287 & 0.255 & 0.339 \\
    
    & 192 &\textbf{0.240}&\textbf{0.299}& 0.250 & 0.309 &\ICLRRevision{0.248}&\ICLRRevision{0.313}&\ICLRRevision{0.257}&\ICLRRevision{0.324}& 0.250 & 0.309 & {0.246} & {0.304} & \underline{0.241} & \underline{0.302} & 0.414 & 0.492 & 0.290 & 0.364 & 0.249 & 0.309 & 0.284 & 0.362 & 0.269 & 0.328 & 0.281 & 0.340  \\
    
    & 336 &\textbf{0.301}&\textbf{0.307}& 0.312 & 0.349&\ICLRRevision{0.309}&\ICLRRevision{0.351}&\ICLRRevision{0.318}& \ICLRRevision{0.362}& 0.311 & 0.348 & {0.307} & \underline{0.342} & \underline{0.305} & {0.343} & 0.597 & 0.542 & 0.377 & 0.422 & 0.321 & 0.351 & 0.369 & 0.427 & 0.325 & 0.366 & 0.339 & 0.372 \\
    
    & 720 &\textbf{0.401}&\textbf{0.397}& 0.411 & 0.406 &\ICLRRevision{0.406}&\ICLRRevision{0.405}&\ICLRRevision{0.412}&\ICLRRevision{0.416}  & 0.412 & 0.407 & {0.407} & \underline{0.398} & \underline{0.402} & {0.400} & 1.730 & 1.042 & 0.558 & 0.524 & 0.408 & 0.403 & 0.554 & 0.522 & 0.421 & 0.415 & 0.433 & 0.432 \\
    \cmidrule(lr){2-28}

    & Avg &\textbf{0.279}&\textbf{0.314}& 0.288 & 0.332 &\ICLRRevision{0.287}&\ICLRRevision{0.333}&\ICLRRevision{0.293}&\ICLRRevision{0.347}& 0.288 & 0.332 & {0.286} & {0.327} & \underline{0.281} & \underline{0.326} & 0.757 & 0.610 & 0.358 & 0.404 & 0.291 & 0.333 & 0.350 & 0.401 & 0.305 & 0.349 & 0.327 & 0.371\\
    
    \midrule
    \multirow{5}{*}{\rotatebox{90}{ETTh1}} 
    &  96 &\textbf{0.374}&\textbf{0.393}& 0.386& 0.405 &\ICLRRevision{0.390}&\ICLRRevision{0.403}&\ICLRRevision{0.398}&\ICLRRevision{0.416}  & 0.386 & 0.405 & 0.386 &  \underline{0.395} & 0.414 & 0.419 & 0.423 & 0.448 & 0.479 & 0.464 &  {0.384} & 0.402 & 0.386 &  {0.400} &  \underline{0.376} & 0.419 & 0.449 & 0.459  \\
    
    & 192 &\underline{0.427}&\textbf{0.422}&0.443 &0.437 &\ICLRRevision{0.451}&\ICLRRevision{0.438}&\ICLRRevision{0.451}&\ICLRRevision{0.446} & 0.441 & 0.436 & 0.437 &  \underline{0.424} & 0.460 & 0.445 & 0.471 & 0.474 & 0.525 & 0.492 &  {0.436} & {0.429} & 0.437 & 0.432 &  \textbf{0.420} & 0.448 & 0.500 & 0.482  \\
    
    & 336 &\textbf{0.447}&\textbf{0.441}& 0.489 &0.468 &\ICLRRevision{0.496}&\ICLRRevision{0.458}&\ICLRRevision{0.497}&\ICLRRevision{0.473} & 0.487 &  {0.458} &  {0.479} &  \underline{0.446} & 0.501 & 0.466 & 0.570 & 0.546 & 0.565 & 0.515 & 0.491 & 0.469 & 0.481 & 0.459 &  \underline{0.459} & 0.465 & 0.521 & 0.496 \\
    
    & 720 &\textbf{0.469}&\textbf{0.463}& 0.502 &0.489 &\ICLRRevision{0.520}&\ICLRRevision{0.493}&\ICLRRevision{0.526}&\ICLRRevision{0.509}& 0.503 & 0.491 &  \underline{0.481} & \underline{0.470} &  {0.500} &  {0.488} & 0.653 & 0.621 & 0.594 & 0.558 & 0.521 & 0.500 & 0.519 & 0.516 & 0.506 & 0.507 & 0.514 & 0.512  \\
    \cmidrule(lr){2-28}
     & Avg &\textbf{0.434}&\textbf{0.430}&  0.455& 0.450 &\ICLRRevision{0.439}&\ICLRRevision{0.448}&\ICLRRevision{0.468}&\ICLRRevision{0.461} & 0.454 &  {0.447} &  \underline{0.446} &  \underline{0.434} & 0.469 & 0.454 & 0.529 & 0.522 & 0.541 & 0.507 & 0.458 & 0.450 & 0.456 & 0.452 &  {0.440} & 0.460 & 0.496 & 0.487\\
     \midrule
    \multirow{5}{*}{\rotatebox{90}{ETTh2}} 
    & 96  &\textbf{0.287}&\textbf{0.337}&  0.296 &0.348 &\ICLRRevision{0.298}&\ICLRRevision{0.351}&\ICLRRevision{0.307}&\ICLRRevision{0.363}  & 0.297 & 0.349 & \underline{0.288} & \underline{0.338} & 0.302 & 0.348 & 0.745 & 0.584 & 0.400 & 0.440 & 0.340 & 0.374 & 0.333 & 0.387 & 0.358 & 0.397 & 0.346 & 0.388 \\
    
    & 192 &\textbf{0.370}&\textbf{0.388}&  0.376 &0.396 &\ICLRRevision{0.393}&\ICLRRevision{0.407}&\ICLRRevision{0.394}&\ICLRRevision{0.414} & 0.380 & 0.400 & \underline{0.374} & \underline{0.390} & 0.388 & 0.400 & 0.877 & 0.656 & 0.528 & 0.509 & 0.402 & 0.414 & 0.477 & 0.476 & 0.429 & 0.439 & 0.456 & 0.452  \\
    
    & 336 &\textbf{0.412}&\textbf{0.425}& 0.424 &{0.431} &\ICLRRevision{0.436}&\ICLRRevision{0.441}&\ICLRRevision{0.437}&\ICLRRevision{0.447} & 0.428 & 0.432 &  \underline{0.415} &  \underline{0.426} & 0.426 & 0.433 & 1.043 & 0.731 & 0.643 & 0.571 & 0.452 & 0.452 & 0.594 & 0.541 & 0.496 & 0.487 & 0.482 & 0.486  \\
    
    & 720 &\textbf{0.419}&\textbf{0.438}& 0.426 &0.444 &\ICLRRevision{0.431}&\ICLRRevision{0.449}&\ICLRRevision{0.445}&\ICLRRevision{0.462}  & 0.427 & 0.445 &  \underline{0.420} &  \underline{0.440} & 0.431 & 0.446 & 1.104 & 0.763 & 0.874 & 0.679 & 0.462 & 0.468 & 0.831 & 0.657 & 0.463 & 0.474 & 0.515 & 0.511 \\
    \cmidrule(lr){2-28}
    & Avg &\textbf{0.372}&\textbf{0.396}& 0.381& 0.405 &\ICLRRevision{0.390}&\ICLRRevision{0.412}&\ICLRRevision{0.396}&\ICLRRevision{0.422}& 0.383 & 0.407 &  \underline{0.374} &  \underline{0.398} & 0.387 & 0.407 & 0.942 & 0.684 & 0.611 & 0.550 & 0.414 & 0.427 & 0.559 & 0.515 & 0.437 & 0.449 & 0.450 & 0.459\\
    \midrule
    \multirow{5}{*}{\rotatebox{90}{Electricity}} 
    & 96  &\textbf{0.137}&\textbf{0.234}& \underline{0.139} &\underline{0.235} &\ICLRRevision{0.192}&\ICLRRevision{0.280}&\ICLRRevision{0.146}&\ICLRRevision{0.246}  &  {0.148} & {0.240} & 0.201 & 0.281 & 0.181 & 0.270 & 0.219 & 0.314 & 0.237 & 0.329 & 0.168 & 0.272 & 0.197 & 0.282 & 0.193 & 0.308 & 0.201 & 0.317\\
    
    & 192 &\textbf{0.158}&\textbf{0.251}&  \underline{0.159} &\underline{0.255}&\ICLRRevision{0.191}&\ICLRRevision{0.280}& \ICLRRevision{0.167}&\ICLRRevision{0.265} &  {0.162} &  {0.253} & 0.201 & 0.283 & 0.188 & 0.274 & 0.231 & 0.322 & 0.236 & 0.330 & 0.184 & 0.289 & 0.196 & 0.285 & 0.201 & 0.315 & 0.222 & 0.334 \\
    
    & 336 &0.182&\textbf{0.173}& \textbf{0.176} &\underline{0.272}&\ICLRRevision{0.211}&\ICLRRevision{0.299}&\ICLRRevision{0.182}&\ICLRRevision{0.281}&  \underline{0.178} &  {0.269} & 0.215 & 0.298 & 0.204 & 0.293 & 0.246 & 0.337 & 0.249 & 0.344 & 0.198 & 0.300 & 0.209 & 0.301 & 0.214 & 0.329 & 0.231 & 0.338\\
    
    & 720 &\textbf{0.200}&\textbf{0.292}&\underline{ 0.204} &\underline{0.298} &\ICLRRevision{0.264}&\ICLRRevision{0.340}&\ICLRRevision{0.208}&\ICLRRevision{0.304}  & 0.225 &  {0.317} & 0.257 & 0.331 & 0.246 & 0.324 & 0.280 & 0.363 & 0.284 & 0.373 &  {0.220} & 0.320 & 0.245 & 0.333 & 0.246 & 0.355 & 0.254 & 0.361\\
    \cmidrule(lr){2-28}
    
    & Avg &\textbf{0.170}&\textbf{0.238}&\textbf{0.170} &\underline{0.265} &\ICLRRevision{0.215}&\ICLRRevision{0.300}&\ICLRRevision{0.176}&\ICLRRevision{0.274} &  {0.178} &  {0.270} & 0.219 & 0.298 & 0.205 & 0.290 & 0.244 & 0.334 & 0.251 & 0.344 & 0.192 & 0.295 & 0.212 & 0.300 & 0.214 & 0.327 & 0.227 & 0.338 \\
    \midrule
    
    \multirow{5}{*}{\rotatebox{90}{Exchange}} 
    & 96  &\textbf{0.085}&\textbf{0.205}& \underline{0.086} &0.207 &\ICLRRevision{0.091}&\ICLRRevision{0.216}&\ICLRRevision{0.103}&\ICLRRevision{0.233} &  {0.086} & \underline{0.206} & 0.093 & 0.217 &  {0.088} &  {0.205} & 0.256 & 0.367 & 0.094 & 0.218 & 0.107 & 0.234 &  {0.088} & 0.218 & 0.148 & 0.278 & 0.197 & 0.323\\
    
    & 192 &\textbf{0.175}&\textbf{0.297}& 0.182 &0.304 &\ICLRRevision{0.189}&\ICLRRevision{0.313}&\ICLRRevision{0.214}&\ICLRRevision{0.337}&  {0.177} &  \underline{0.299} & 0.184 & 0.307 &  \underline{0.176} &  {0.299} & 0.470 & 0.509 & 0.184 & 0.307 & 0.226 & 0.344 &  {0.176} & 0.315 & 0.271 & 0.315 & 0.300 & 0.369\\
    
    & 336 &{0.317}&\underline{0.407}& 0.332 &0.418 &\ICLRRevision{0.333}&\ICLRRevision{0.421}&\ICLRRevision{0.366}&\ICLRRevision{0.445} & 0.331 &  {0.417} & 0.351 & 0.432 &  \textbf{0.301} &  \textbf{0.397} & 1.268 & 0.883 & 0.349 & 0.431 & 0.367 & 0.448 &  \underline{0.313} & 0.427 & 0.460 & 0.427 & 0.509 & 0.524\\
    
    & 720 &\textbf{0.825}&\textbf{0.683}& 0.867 &0.703&\ICLRRevision{0.916}&\ICLRRevision{0.729}&\ICLRRevision{0.931}&\ICLRRevision{0.738} &  {0.847} &  \underline{0.691} & 0.886 & 0.714 & 0.901 & 0.714 & 1.767 & 1.068 & 0.852 & 0.698 & 0.964 & 0.746 &  \underline{0.839} &  {0.695} & 1.195 &  {0.695} & 1.447 & 0.941\\

    \cmidrule(lr){2-28}
    & Avg &\textbf{0.351}&\textbf{0.400}& 0.367 &0.408&\ICLRRevision{0.382}&\ICLRRevision{0.420}&\ICLRRevision{0.404}&\ICLRRevision{0.428} &  \underline{0.360} &  \underline{0.403} & 0.378 & 0.417 & 0.367 &  {0.404} & 0.940 & 0.707 & 0.370 & 0.413 & 0.416 & 0.443 &  {0.354} & 0.414 & 0.519 & 0.429 & 0.613 & 0.539\\
    \midrule
    \multirow{5}{*}{\rotatebox{90}{Solar-Energy}} 
    & 96  &\textbf{0.202}&\textbf{0.233}&  0.205 & 0.244&\ICLRRevision{0.238}&\ICLRRevision{0.277}&\ICLRRevision{0.231}&\ICLRRevision{0.286} &  \underline{0.203} &  \underline{0.237} & 0.322 & 0.339 & 0.234 & 0.286 & 0.310 & 0.331 & 0.312 & 0.399 & 0.250 & 0.292 & 0.290 & 0.378 & 0.242 & 0.342 & 0.884 & 0.711  \\
    
    & 192 &\textbf{0.230}&\textbf{0.254}& 0.237 &0.270&\ICLRRevision{0.299}&\ICLRRevision{0.319}&\ICLRRevision{0.257}&\ICLRRevision{0.285}&  \underline{0.233} &  \underline{0.261} & 0.359 & 0.356 & 0.267 & 0.310 & 0.734 & 0.725 & 0.339 & 0.416 & 0.296 & 0.318 & 0.320 & 0.398 & 0.285 & 0.380 & 0.834 & 0.692  \\
    
    & 336 &\underline{0.254}&\textbf{0.265}& 0.258 &0.288&\ICLRRevision{0.310}&\ICLRRevision{0.327}&\ICLRRevision{0.256}&\ICLRRevision{0.293}  &  \textbf{0.248} & \underline{0.273} & 0.397 & 0.369 & 0.290 & 0.315 & 0.750 & 0.735 & 0.368 & 0.430 & 0.319 & 0.330 & 0.353 & 0.415 & 0.282 & 0.376 & 0.941 & 0.723 \\
    
    & 720 &\underline{0.252}&\textbf{0.271}& 0.260 &0.288&\ICLRRevision{0.310}&\ICLRRevision{0.330}&\ICLRRevision{0.252}&\ICLRRevision{0.295}  &  \textbf{0.249} & \underline{0.275} & 0.397 & 0.356 & 0.289 & 0.317 & 0.769 & 0.765 & 0.370 & 0.425 & 0.338 & 0.337 & 0.356 & 0.413 & 0.357 & 0.427 & 0.882 & 0.717 \\
    \cmidrule(lr){2-28}
    
    & Avg &\underline{0.235}&\textbf{0.256}&0.240 &0.273&\ICLRRevision{0.289}&\ICLRRevision{0.313}&\ICLRRevision{0.249}&\ICLRRevision{0.290}&  \textbf{0.233} &  \underline{0.262} & 0.369 & 0.356 & 0.270 & 0.307 & 0.641 & 0.639 & 0.347 & 0.417 & 0.301 & 0.319 & 0.330 & 0.401 & 0.291 & 0.381 & 0.885 & 0.711 \\\midrule
    \multirow{5}{*}{\rotatebox{90}{PEMS04}} 
    & 12 &\textbf{0.075}& \underline{0.182}&  \underline{0.076} &  \textbf{0.180} &\ICLRRevision{0.110}&\ICLRRevision{0.226}&\ICLRRevision{0.082}&\ICLRRevision{0.193}&  {0.078} &  {0.183} & 0.138 & 0.252 & 0.105 & 0.224 & 0.098 & 0.218 & 0.219 & 0.340 & 0.087 & 0.195 & 0.148 & 0.272 & 0.138 & 0.262 & 0.424 & 0.491 \\
    
    & 24 &\textbf{0.084}&\textbf{0.193}& \textbf{0.084} &  \textbf{0.193}&\ICLRRevision{0.161}&\ICLRRevision{0.275}&\ICLRRevision{0.099}&\ICLRRevision{0.214} &  {0.095} &  {0.205} & 0.258 & 0.348 & 0.153 & 0.275 & 0.131 & 0.256 & 0.292 & 0.398 & 0.103 & 0.215 & 0.224 & 0.340 & 0.177 & 0.293 & 0.459 & 0.509\\
    
    & 48 &\textbf{0.105}&\textbf{0.217}&  \underline{0.115} &  \underline{0.224} &\ICLRRevision{0.345}&\ICLRRevision{0.403}&\ICLRRevision{0.123}&\ICLRRevision{0.240} &  {0.120} &  {0.233} & 0.572 & 0.544 & 0.229 & 0.339 & 0.205 & 0.326 & 0.409 & 0.478 & 0.136 & 0.250 & 0.355 & 0.437 & 0.270 & 0.368 & 0.646 & 0.610\\
    
    & 96 &\textbf{0.130}&\textbf{0.243}& \underline{0.137} &  \underline{0.248}&\ICLRRevision{0.588}&\ICLRRevision{0.553}&\ICLRRevision{0.151}&\ICLRRevision{0.267} &  {0.150} &  {0.262} & 1.137 & 0.820 & 0.291 & 0.389 & 0.402 & 0.457 & 0.492 & 0.532 & 0.190 & 0.303 & 0.452 & 0.504 & 0.341 & 0.427 & 0.912 & 0.748\\
    \cmidrule(lr){2-28}
    
    & Avg &\textbf{0.099}&\textbf{0.209}& \underline{0.103} & \underline{0.211}&\ICLRRevision{0.301}&\ICLRRevision{0.364}&\ICLRRevision{0.114}&\ICLRRevision{0.229} &  {0.111} &  {0.221} & 0.526 & 0.491 & 0.195 & 0.307 & 0.209 & 0.314 & 0.353 & 0.437 & 0.129 & 0.241 & 0.295 & 0.388 & 0.231 & 0.337 & 0.610 & 0.590\\\midrule
    
    \multirow{5}{*}{\rotatebox{90}{PEMS08}}
    & 12 &\textbf{0.075}&\textbf{0.177}& \underline{0.076} &  \underline{0.178} &\ICLRRevision{0.099}&\ICLRRevision{0.214}&\ICLRRevision{0.080}&\ICLRRevision{0.190}&  {0.079} &  {0.182} & 0.133 & 0.247 & 0.168 & 0.232 & 0.165 & 0.214 & 0.227 & 0.343 & 0.112 & 0.212 & 0.154 & 0.276 & 0.173 & 0.273 & 0.436 & 0.485  \\
    
    & 24 &\textbf{0.102}&\textbf{0.207}& \underline{0.104} &  \underline{0.209}&\ICLRRevision{0.169}&\ICLRRevision{0.277}&\ICLRRevision{0.114}&\ICLRRevision{0.223} &  {0.115} &  {0.219} & 0.249 & 0.343 & 0.224 & 0.281 & 0.215 & 0.260 & 0.318 & 0.409 & 0.141 & 0.238 & 0.248 & 0.353 & 0.210 & 0.301 & 0.467 & 0.502  \\
    
    & 48 &\textbf{0.154}&\textbf{0.226} &\underline{0.167} &  \underline{0.228} &\ICLRRevision{0.274}&\ICLRRevision{0.360}&\ICLRRevision{0.175}&\ICLRRevision{0.271} &  {0.186} &  {0.235} & 0.569 & 0.544 & 0.321 & 0.354 & 0.315 & 0.355 & 0.497 & 0.510 & 0.198 & 0.283 & 0.440 & 0.470 & 0.320 & 0.394 & 0.966 & 0.733 \\
    
    & 96 &\underline{0.243}&0.305& 0.245 & \underline{0.280} &\ICLRRevision{0.522}&\ICLRRevision{0.499}&\ICLRRevision{0.298}&\ICLRRevision{0.348}&\textbf{0.221} & \textbf{0.267} & 1.166 & 0.814 & 0.408 & 0.417 & 0.377 & 0.397 & 0.721 & 0.592 & 0.320 & 0.351 & 0.674 & 0.565 & 0.442 & 0.465 & 1.385 & 0.915  \\
    \cmidrule(lr){2-28}
    
    & Avg &\textbf{0.145}&0.228 &  \underline{0.148} &  \textbf{0.223}&\ICLRRevision{0.266}&\ICLRRevision{0.338}&\ICLRRevision{0.167}&\ICLRRevision{0.258} &  {0.150} &  \underline{0.226} & 0.529 & 0.487 & 0.280 & 0.321 & 0.268 & 0.307 & 0.441 & 0.464 & 0.193 & 0.271 & 0.379 & 0.416 & 0.286 & 0.358 & 0.814 & 0.659\\
    
    \bottomrule
  \end{tabular}
  \end{small}
}
\vspace{-0.1in}
\end{table*}

\section{Evaluation}
\label{sec:eval}

In this section, we aim to conduct experiments to answer the following questions: \textbf{Q1:} What is the effectiveness of \model\ compared with other state-of-the-art baselines? \textbf{Q2:} How each component of \model\  affect the final performance? \textbf{Q3:} How is the robustness of \model\ compared with state-of-the-art methods like S-Mamba and iTransfomrer? \textbf{Q4:} How is the advantage of \model\ on long-term prediction with increasing lookback length compared to other state-of-the-art methods? \textbf{Q5:} How is the performance of \model\ on capturing multi-scale periodicity and transient dynamics compared to state-of-the-art baselines? \textbf{Q6:} How is the efficiency of \model\ compared to state-of-the-art baselines? \textbf{Q7:} How do hyperparameters of \model affect the performance? \textbf{Q8:} How is \model\ affected by long Lookback comparison compared with several state-of-the-art baselines like S-Mamba, iTransfomrer?

\subsection{Experimental Setup}
\label{sec:setup}
\textbf{Datasets}. To rigorously evaluate the effectiveness of our proposed model, we selected a diverse set of 9 real-world datasets~\citep{haoyietal-informerEx-2023,haoyietal-informer-2021} for evaluation. These datasets encompass a range of domains, including Electricity, 4 ETT datasets (ETTh1, ETTh2, ETTm1, ETTm2), and others. These datasets are extensively utilized in research and span various fields, such as transportation analysis and energy management. Detailed statistics for each dataset can be found in Table ~\ref{tab:datasets}. 

\begin{table}[htbp]
\vspace{-0.1in}
\begin{center}
\renewcommand{\arraystretch}{0.7}
\caption{The statistics of 9 public datasets.}
\label{tab:datasets}
    \begin{tabular}{c|ccc}
    \toprule
    Datasets & Variates & Timesteps & Granularity  \\
    \hline
    ETTh1\&ETTh2 & 7 & 69,680 & 1 hour  \\
    PEMS04 & 307 & 16,992 & 5 minutes  \\
    PEMS08 & 170 & 17,856 & 5 minutes  \\ 
    Exchange & 8 &  7,588 &  1 day\\
    Electricity & 321 & 26,304 & 1 hour\\
    Solar-Energy & 137 & 52,560 & 10 minutes\\
    ETTm1\&ETTm2 & 7 & 17,420 & 15min\\
    \bottomrule
    \end{tabular}
\end{center}
\vspace{-0.7cm}
\end{table}

\textbf{Experiment Settings}.
To ensure a fair comparison, we modify the hidden dimensionality of all compared algorithms within the range of $[128, 256, 512, 1024, 2048]$ to achieve their reported best performance, which is consistently observed at 1024. The learning rate ($\eta$) is initialized to $5\times10^{-6}$, and we set the number of \model\ layers to 2. Consistent with the existing settings of time series datasets, we utilize historical data with 96, 192, 336, or 720 time steps. The time steps are defined as 5 minutes, 1 hour, 10 minutes, or 1 day intervals to predict the corresponding future 96, 192, 336, or 720 time steps in these time series datasets. All baseline methods are evaluated using their predefined settings as described in their respective publications. We conduct testing for all tasks on a single NVIDIA L40 GPU equipped with 128 CPUs.

\textbf{Baseline Descriptions}. We compare our method \model\ with 12 state-of-the-art methods including 6 Transformer-based models, 3 MLP-based methods and 3 SSM-based method. The detailed illustrations are shown as follows:
\underline{Transformer-based methods:} 
\begin{itemize}
    \item Autoformer~\citep{wu2021autoformer} employs a series decomposition technique along with an Auto-Correlation mechanism to effectively capture cross-time dependencies.
    \item FEDformer~\citep{zhou2022fedformer} introduces an enhanced Transformer operating in the frequency domain, aiming to improve both efficiency and effectiveness.
    \item Crossformer~\citep{zhang2022crossformer} incorporates a patching operation like other models but distinguishes itself by employing Cross-Dimension attention to capture dependencies between different series. While patching reduces the elements to process and extracts semantic information comprehensively, these models encounter performance limitations when handling longer. 
    \item DLinear~\citep{zeng2023transformers} introduced DLinear, a method that decomposes time series into two distinct components and generates a single Linear layer for each component. This straightforward design has outperformed all previously proposed complex transformer models.
    \item PatchTST~\citep{huang2024long} leverages patching and channel-independent techniques to facilitate the extraction of semantic information from single time steps to multiple time steps within time series data.
    \item iTransFormer~\citep{liu2023itransformer} employs inverted attention layers to effectively capture inter-series dependencies. However, its tokenization approach, which involves passing the entire sequence through a Multilayer Perceptron (MLP) layer, falls short in capturing the complex evolutionary patterns inherent in time series data.
\end{itemize}

\begin{figure}
\vspace*{-4mm}
\centering
\begin{tabular}{c c }
\\\hspace{-4.0mm}
  \begin{minipage}{0.20\textwidth}
	\includegraphics[width=\textwidth]{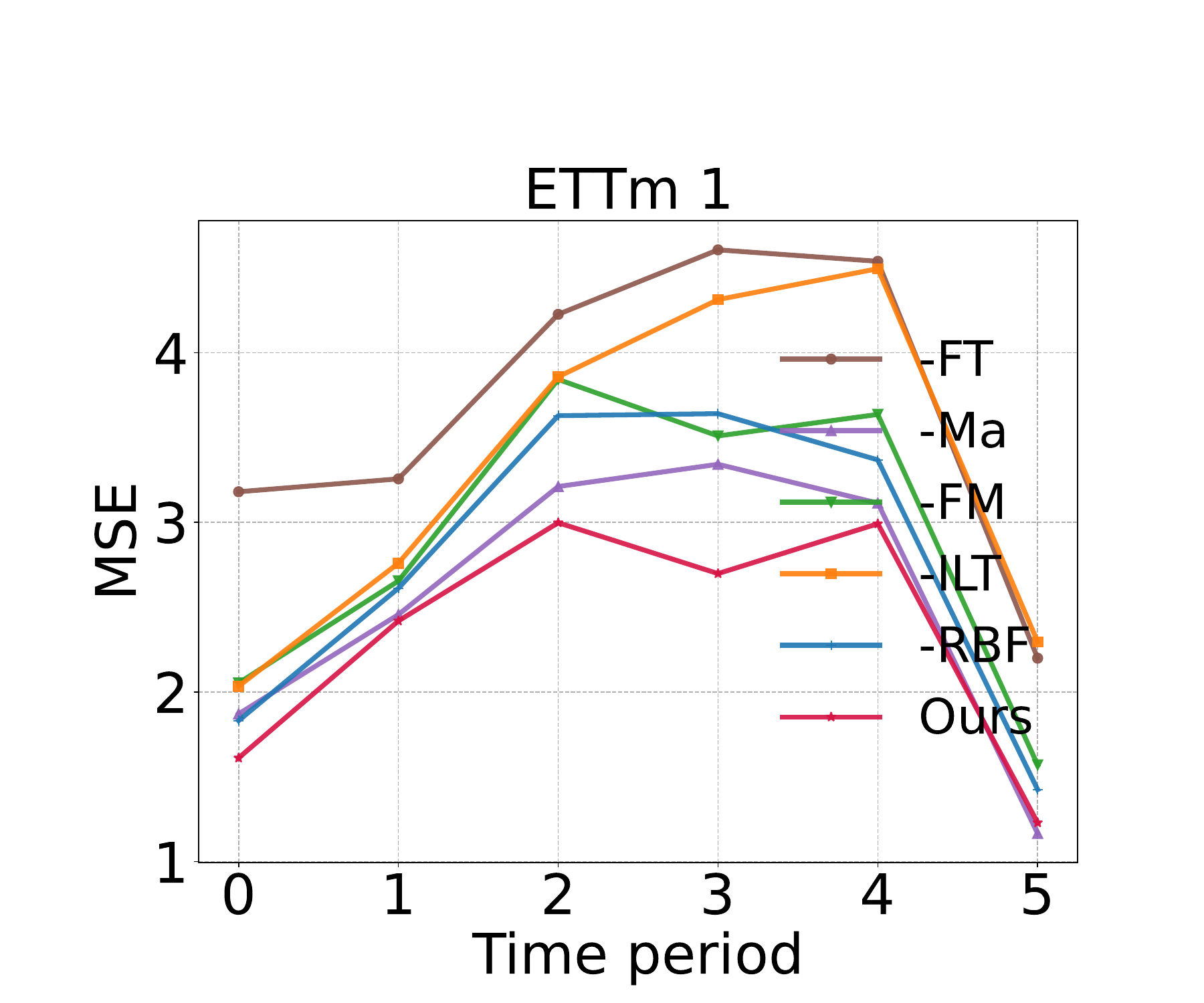}
  \end{minipage}\hspace{-3.mm}
  &
  \begin{minipage}{0.20\textwidth}
	\includegraphics[width=\textwidth]{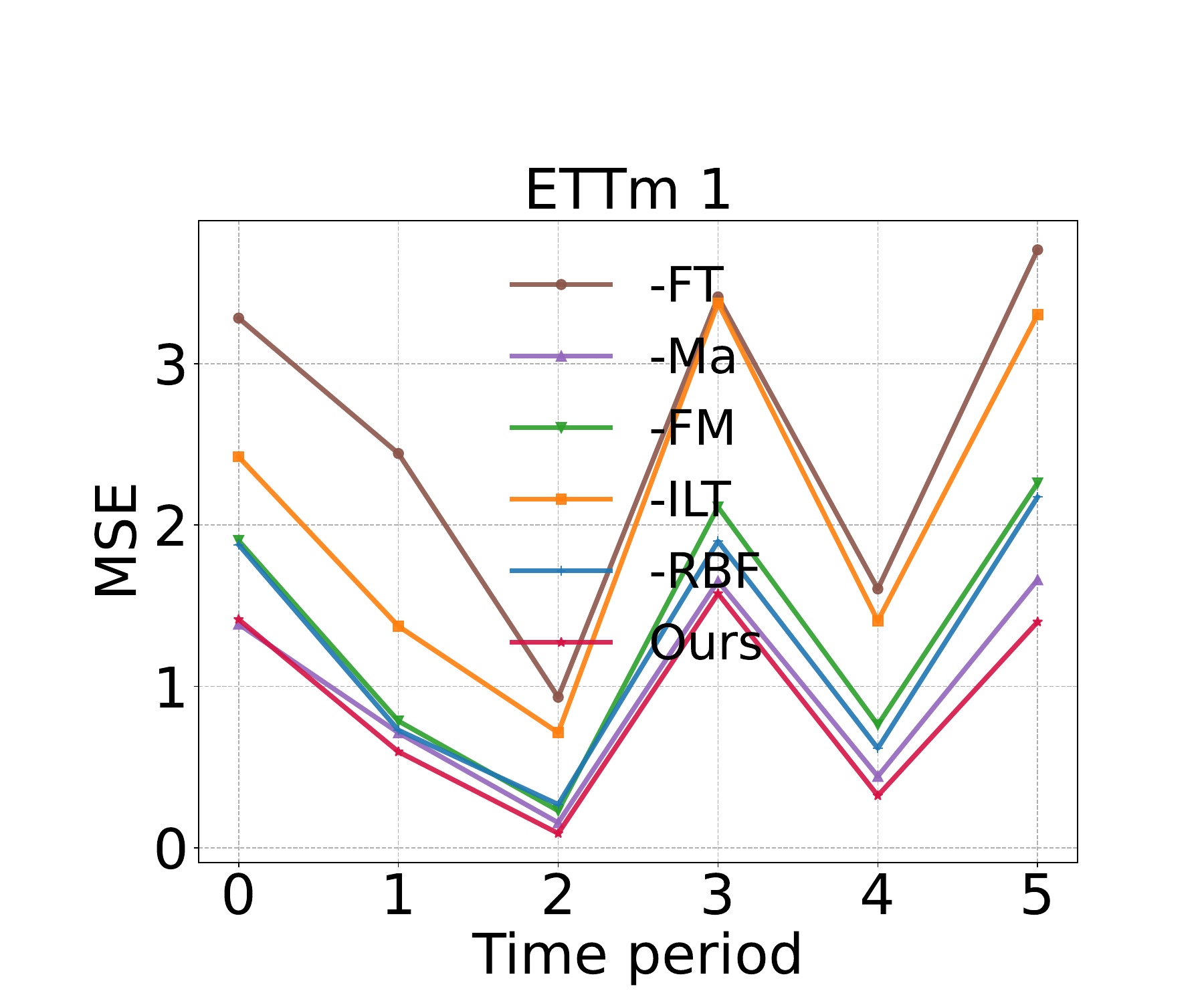}
  \end{minipage}
\\\hspace{-4.0mm}
(a) Node 9\hspace{2.0mm}
& (b) Node 18\hspace{-4.0mm}
\end{tabular}
\caption{Ablation study of \model on prediction performance on Node 9 and Node 18 instances of ETTm 1 dataset.}
\label{fig:ablation_pre}
\vspace*{-1mm}
\end{figure}

\underline{MLP-based methods:} 
\begin{itemize}
    \item TimesNet~\citep{wu2022timesnet} expands the examination of temporal fluctuations by extending the 1-D time series into a collection of 2-D tensors across multiple periods. 
    \item RLinear~\citep{li2023revisiting}, the state-of-the-art linear model, incorporates reversible normalization and channel independence into a purely linear structure.
    \item TiDE~\citep{das2023long} is an encoder-decoder model that employs a Multi-layer Perceptron (MLP) architecture. 
\end{itemize}

\underline{SSM-based methods:}
\begin{itemize}
    \item S-Mamba~\citep{wang2025mamba} independently tokenizes the time points for each variate using a linear layer. This allows for the extraction of correlations between variates using a bidirectional Mamba layer, while a Feed-Forward Network is employed to learn temporal dependencies.
    \item SST~\citep{xusst} leverages Mamba to identify global patterns in coarse-grained long-range time series, while the Local Window Transformer (LWT) focuses on local variations in fine-grained short-range time series.
    \item Bi-mamba+~\cite{liang2024bi} introduces a patching technique aimed at enhancing local information and capturing evolving patterns to address sparse time series semantics, primarily targeting long-term predictions with high efficiency.
\end{itemize}

\subsection{Overall Comparison (Q1)}
We present evaluation results via two metrics: Mean Squared Error (MSE) and Mean Absolute Error (MAE) (Table~\ref{tab:results_other}). We make the following observations:

\textbf{Outstanding Performance}. Our proposed framework, \model, demonstrates exceptional performance across a range of time series prediction tasks.  As shown in Table~\ref{tab:results_other}, \model achieves state-of-the-art results in the majority of scenarios (60 out of 72, or 83.3\%), and consistently ranks among the top performers in the remaining cases across nine real-world datasets. This outstanding performance can be attributed to several key design elements: \textbf{(1) Data Smoothing via the Radial Basis Function (RBF) Kernel:} \model incorporates an RBF kernel, which effectively smooths the input data, reducing noise and enabling more accurate capture of underlying temporal patterns. This data preprocessing step significantly contributes to the model's improved prediction accuracy. \textbf{(2) Multi-Scale Periodicity Capture with the Fast Fourier Transform (FFT):}  Our framework incorporates the FFT on the parameter $\Delta$. This transformation enables the identification and extraction of multi-scale periodic patterns present in the time series data. By effectively capturing these periodic patterns, \model significantly enhances its predictive capabilities. \textbf{(3) Enhanced Long-Term Prediction and Transient Dynamics Capture with the Inverse Laplace Transform:} To further improve long-term predictions and capture transient dynamics, \model incorporates the inverse Laplace transform on the combined outputs of \smodel and Mamba. This innovative approach proves advantageous in capturing both transient dynamics and periodic patterns, further boosting the accuracy of our prediction outputs. {\textbf{(4) Integration of \smodel and Mamba via the \module Block:}} The \module block within \model facilitates the capture of complex temporal attributes and dependencies between the \smodel and Mamba components. This integration enhances the model's ability to capture intricate temporal relationships, improving overall performance.

\textbf{Pearson Correlation Comparison}.
We also calculated Pearson correlation and show results in Table~\ref{tab:results_other_horizontal}. The results indicate that our method consistently outperforms other baselines across most cases and all datasets, further confirming its superior performance.
The superior performance \model in time series prediction arises from the synergistic combination of an RBF kernel, Fast Fourier and inverse Laplace transforms, and the integrated \smodel and Mamba components. This approach effectively captures temporal patterns, multi-scale periodicity, transient dynamics, and mitigates noise.

\begin{figure}
\vspace*{-4mm}
\centering
\begin{tabular}{c c}
\\\hspace{-1.0mm}
  \begin{minipage}{0.22\textwidth}
	\includegraphics[width=\textwidth]{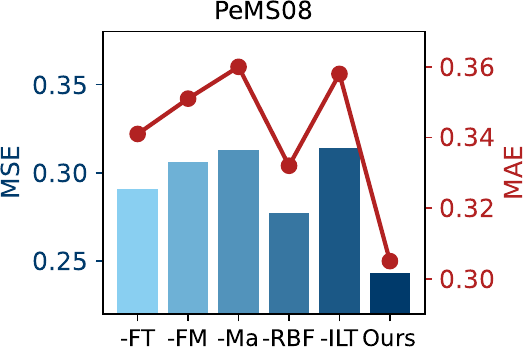}
  \end{minipage}\hspace{-3.0mm}
  &
  \begin{minipage}{0.22\textwidth}
	\includegraphics[width=\textwidth]{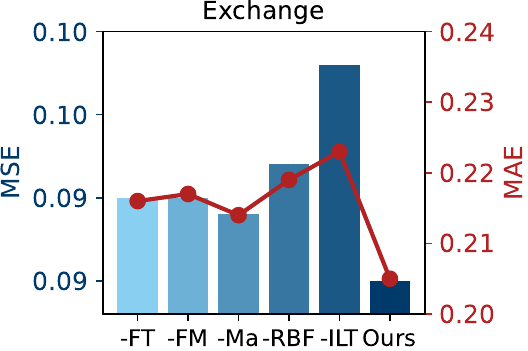}
  \end{minipage}\hspace{-3.0mm}
  \\
  \begin{minipage}{0.22\textwidth}
	\includegraphics[width=\textwidth]{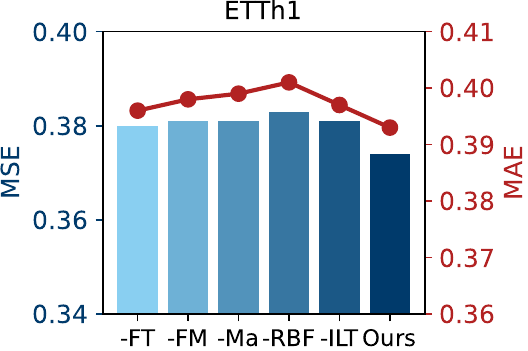}
  \end{minipage}\hspace{-3.0mm}
  &
  \begin{minipage}{0.22\textwidth}
	\includegraphics[width=\textwidth]{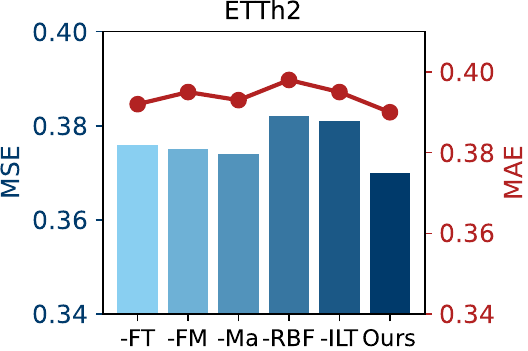}
  \end{minipage}\hspace{-3.0mm}
\end{tabular}
\caption{Ablation study of \model with $L = 96$.}
\vspace*{-0.1in}
\label{fig:ablation}
\end{figure}

\begin{table*}[htb]
\vspace{-0.1in}
\caption{{\ICLRRevision{Performance comparison in terms of Pearson correlation}}}
\label{tab:results_other_horizontal}
\renewcommand{\arraystretch}{1.0}
\centering
\resizebox{0.8\textwidth}{!}{ 
\begin{small}
\setlength{\tabcolsep}{2.6pt}
\vspace{1mm}
\begin{tabular}{c|c|c|c|c|c|c|c|c|c|c|c}
    \toprule
    \multicolumn{1}{c|}{{\ICLRRevision{Models}}} & \multicolumn{1}{c|}{{\ICLRRevision{Metric}}} & \multicolumn{1}{c|}{{\ICLRRevision{ETTm1 }}} & \multicolumn{1}{c|}{{\ICLRRevision{ETTm2 }}} & \multicolumn{1}{c|}{{\ICLRRevision{ETTh1 }}} & \multicolumn{1}{c|}{{\ICLRRevision{ETTh2 }}} & \multicolumn{1}{c|}{{\ICLRRevision{Electricity}}} & \multicolumn{1}{c|}{{\ICLRRevision{Exchange}}} & \multicolumn{1}{c|}{{\ICLRRevision{Solar-Energy}}} & \multicolumn{1}{c|}{{\ICLRRevision{Metric}}} & \multicolumn{1}{c|}{{\ICLRRevision{PEMS04}}}& \multicolumn{1}{c}{{\ICLRRevision{PEMS08}}}\\
    \toprule
    \multirow{5}{*}{{\ICLRRevision{\model(\textbf{ours)}}}} & 96  & {\ICLRRevision{\textbf{0.857}}} & {\ICLRRevision{\textbf{0.950}}} & {\ICLRRevision{\textbf{0.892}}} & {\ICLRRevision{\textbf{0.920}}} & {\ICLRRevision{\underline{0.929}}} & {\ICLRRevision{\textbf{0.978}}} & {\ICLRRevision{\textbf{0.818}}} & 12 & {\ICLRRevision{\textbf{0.793}}}&{\ICLRRevision{\textbf{0.839}}}\\
    
    & 192 & {\ICLRRevision{\textbf{0.830}}} & {\ICLRRevision{\textbf{0.935}}} & {\ICLRRevision{\textbf{0.799}}} & {\ICLRRevision{\textbf{0.898}}} & {\ICLRRevision{\textbf{0.920}}} & {\ICLRRevision{\textbf{0.958}}} &{\ICLRRevision{\textbf{0.856}}}& 24 & {\ICLRRevision{\textbf{0.768}}} &{\ICLRRevision{\textbf{0.802}}}\\
    
    & 336 & {\ICLRRevision{\textbf{0.812}}} & {\ICLRRevision{\textbf{0.920}}} & {\ICLRRevision{\textbf{0.776}}} & {\ICLRRevision{\textbf{0.882}}} & {\ICLRRevision{\textbf{0.912}}} & {\ICLRRevision{\textbf{0.926}}} &{\ICLRRevision{0.839}} &48&{\ICLRRevision{\underline{0.765}}} &{\ICLRRevision{\textbf{0.775}}}\\
    
    & 720 & {\ICLRRevision{\textbf{0.781}}} & {\ICLRRevision{\textbf{0.896}}} & {\ICLRRevision{\textbf{0.766}}} & {\ICLRRevision{\textbf{0.886}}} & {\ICLRRevision{\textbf{0.890}}} & {\ICLRRevision{\textbf{0.844}}} &{\ICLRRevision{0.820}} &96&{\ICLRRevision{\textbf{0.815}}}&{\ICLRRevision{\textbf{0.777}}}\\
    \cmidrule(lr){2-12}
     & Avg & {\ICLRRevision{\textbf{0.820}}} & {\ICLRRevision{\textbf{0.925}}} & {\ICLRRevision{\textbf{0.793}}} & {\ICLRRevision{\textbf{0.897}}} & {\ICLRRevision{\textbf{0.913}}} & {\ICLRRevision{\textbf{0.927}}} &{\ICLRRevision{\textbf{0.833}}} &Avg&{\ICLRRevision{\textbf{0.785}}}&{\ICLRRevision{\textbf{0.798}}}\\
    \midrule
    
    \multirow{5}{*}{{\ICLRRevision{S-Mamba}}} & 96  & {\ICLRRevision{\underline{0.853}}} &{\ICLRRevision{\underline{0.947}}} &{\ICLRRevision{0.825}}  &{\ICLRRevision{\underline{0.909}}}  & {\ICLRRevision{\textbf{0.930}}} & {\ICLRRevision{\underline{0.970}}} &{\ICLRRevision{0.814}} &12&{\ICLRRevision{\underline{0.792}}}&{\ICLRRevision{\underline{0.836}}}\\
    & 192 &  {\ICLRRevision{{0.825}}} &{\ICLRRevision{\underline{0.932}}} & {\ICLRRevision{0.796}} & {\ICLRRevision{\textbf{0.898}}} & {\ICLRRevision{\textbf{0.920}}} & {\ICLRRevision{\underline{0.946}}} &{\ICLRRevision{0.85}}&24&{\ICLRRevision{\underline{0.767}}}&{\ICLRRevision{\underline{0.796}}}\\
    
    & 336 & {\ICLRRevision{\underline{0.808}}} & {\ICLRRevision{\underline{0.916}}} & {\ICLRRevision{0.768}} & {\ICLRRevision{0.874}} & {\ICLRRevision{\underline{0.910}}} & {\ICLRRevision{0.915}} &{\ICLRRevision{\textbf{0.841}}} &48&{\ICLRRevision{\textbf{0.768}}}&{\ICLRRevision{\underline{0.768}}}\\
    
    & 720 & {\ICLRRevision{0.755}} & {\ICLRRevision{\underline{0.895}}} & {\ICLRRevision{\underline{0.756}}} & {\ICLRRevision{0.867}} & {\ICLRRevision{\underline{0.888}}} & {\ICLRRevision{\textbf{0.827}}} &{\ICLRRevision{0.827}}&96&{\ICLRRevision{\underline{0.813}}}&{\ICLRRevision{\underline{0.774}}}\\
    \cmidrule(lr){2-12}
     & Avg & {\ICLRRevision{{0.810}}} & {\ICLRRevision{\underline{0.922}}} & {\ICLRRevision{0.786}} & {\ICLRRevision{\textbf{0.887}}} & {\ICLRRevision{\underline{0.912}}} & {\ICLRRevision{\underline{0.914}}} &{\ICLRRevision{\textbf{0.833}}}&Avg&{\ICLRRevision{\textbf{0.785}}}&{\ICLRRevision{\underline{0.793}}}\\
    \midrule
    \multirow{5}{*}{{\ICLRRevision{iTransformer}}} & 96  & {\ICLRRevision{{0.851}}} & {\ICLRRevision{{0.947}}} & {\ICLRRevision{\underline{0.826}}} & {\ICLRRevision{\underline{0.909}}} & {\ICLRRevision{0.925}} &{\ICLRRevision{\underline{0.970}}} &{\ICLRRevision{\underline{0.816}}} &12&{\ICLRRevision{0.785}}&{\ICLRRevision{0.829}}\\
    
    & 192 & {\ICLRRevision{\underline{0.827}}} & {\ICLRRevision{{0.930}}} & {\ICLRRevision{\textbf{0.799}}} & {\ICLRRevision{0.877}} & {\ICLRRevision{0.918}} & {\ICLRRevision{\underline{0.946}}} & {\ICLRRevision{\underline{0.851}}}&24&{\ICLRRevision{0.748}}&{\ICLRRevision{0.780}}\\
    
    & 336 & {\ICLRRevision{{0.806}}} & {\ICLRRevision{{0.915}}} & {\ICLRRevision{\underline{0.769}}} & {\ICLRRevision{\underline{0.875}}} & {\ICLRRevision{0.910}} & {\ICLRRevision{\underline{0.916}}} & {\ICLRRevision{\underline{0.840}}}&48&{\ICLRRevision{0.733}}&{\ICLRRevision{0.725}}\\
    
    & 720 & {\ICLRRevision{\textbf{0.781}}} & {\ICLRRevision{{0.892}}} & {\ICLRRevision{0.755}} & {\ICLRRevision{\underline{0.869}}} & {\ICLRRevision{{0.887}}} & {\ICLRRevision{{0.826}}} & {\ICLRRevision{\underline{0.821}}}&96&{\ICLRRevision{0.787}}&{\ICLRRevision{0.696}}\\
    
    \cmidrule(lr){2-12}
     & Avg & {\ICLRRevision{\underline{0.816}}} & {\ICLRRevision{{0.921}}} & {\ICLRRevision{\underline{0.787}}} & {\ICLRRevision{0.855}} & {\ICLRRevision{0.910}} & {\ICLRRevision{\underline{0.914}}} &{\ICLRRevision{\underline{0.832}}} &Avg&{\ICLRRevision{0.763}}&{\ICLRRevision{0.757}}\\
    
    \bottomrule
\end{tabular}
\end{small}
}
\vspace{-0.2in}
\end{table*}

\subsection{Ablation Study (Q2)}\label{sec:ablation}
This section aims to evaluate the individual contributions of each component within our proposed framework, \model, as illustrated in Figure~\ref{fig:ablation} and Figure~\ref{fig:ablation_pre}. {\tkde{We conduct an ablation study with five variants: (1) \textbf{``w/o FT'':} removes the Fourier transform's role in $\Delta$, eliminating both forward/inverse transforms for frequency analysis without replacements; (2) \textbf{``w/o FM'':} retains only Mamba to isolate the frequency-enhanced module's contribution; (3) \textbf{``w/o Ma'':} keeps only the frequency-modeling component (\smodel) to assess Mamba's impact; (4) \textbf{``w/o RBF'':} omits the RBF kernel to evaluate data smoothing effects; (5) \textbf{``w/o ILT'':} discards the inverse Laplace transform, leaving data in the frequency domain - compensated via a single-layer MLP for cross-frequency bridging.}}.

\begin{figure}
\vspace*{-3mm}
\centering
\begin{tabular}{c c}
\\\hspace{-4.0mm}
  \begin{minipage}{0.23\textwidth}
	\includegraphics[width=\textwidth]{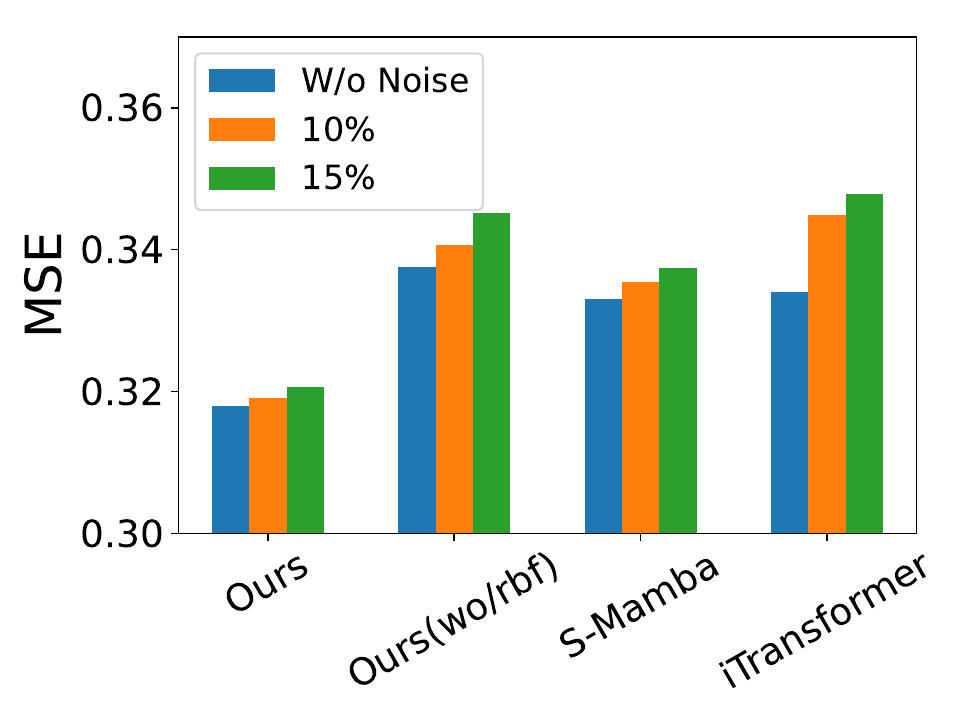}
  \end{minipage}\hspace{-3.mm}
  &
  \begin{minipage}{0.23\textwidth}
	\includegraphics[width=\textwidth]{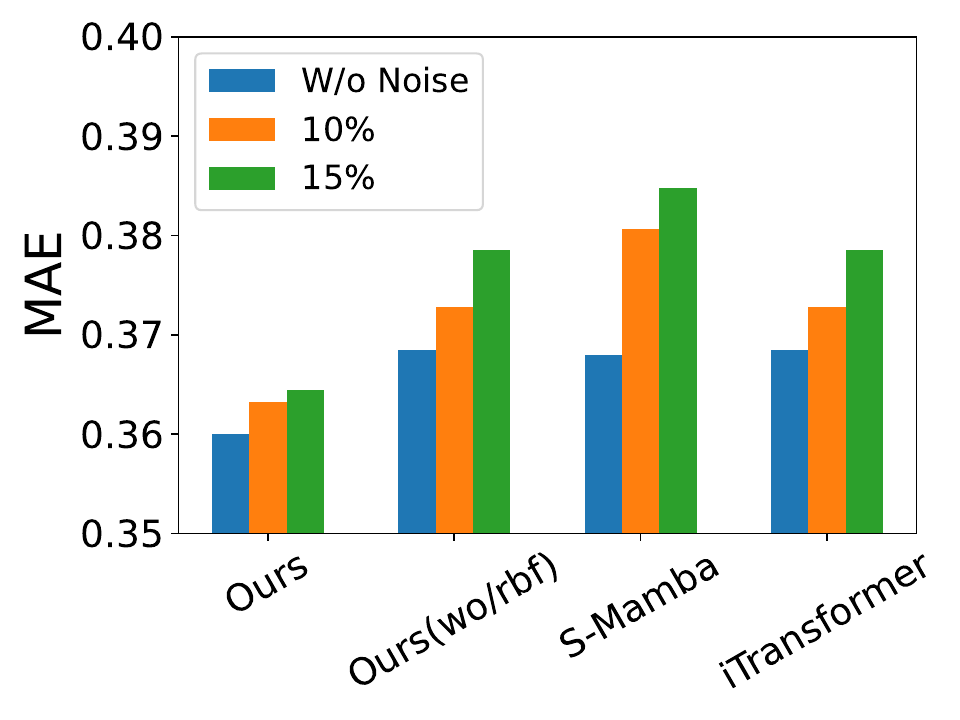}
  \end{minipage}\hspace{-3.0mm}
\end{tabular}
\vspace*{-0.1in}
\caption{Performance comparison of robustness}
\vspace*{-0.1in}
\label{fig:robust}
\end{figure}

By comparing the performance of these variants against our full method, \model, we can isolate the individual contribution of each component to overall performance. The results presented in Figure~\ref{fig:ablation} and Figure~\ref{fig:ablation_pre} demonstrate that each component of \model positively influences performance, confirming the effectiveness of our approach. Notably, the inverse Laplace transform exhibits the most significant impact on the overall effectiveness of our method.

\subsection{Robustness (Q3)} \label{sec:robuseness}
{\tkde{This study evaluates our proposed \model's noise robustness against three baselines: our RBF-kernel-ablated variant, S-Mamba, and iTransformer. Using the ETTm1 dataset with controlled 10\% and 15\% synthetic noise injections, results (Figure~\ref{fig:robust}) demonstrate \model's superior resilience - showing a smaller performance decay than S-Mamba and 2.1× higher noise tolerance than iTransformer. The RBF kernel contributes this robustness (verified through ablation), while the integrated Fourier-Laplace transforms also makes contributions to the MAE improvement over S-Mamba. Notably, \model\ maintains stable prediction quality across noise levels where iTransformer degrades. These findings quantitatively validate: (1) the RBF kernel's noise-filtering efficacy, (2) the dual-transform architecture's advantage in preserving signal fidelity, and (3) \model's architectural superiority for noisy time-series scenarios compared to both pure attention (iTransformer) and single-transform (S-Mamba) approaches.
}}

\begin{figure}
\centering
\begin{tabular}{c c}
\\\hspace{-4.0mm}
  \begin{minipage}{0.21\textwidth}
	\includegraphics[width=\textwidth]{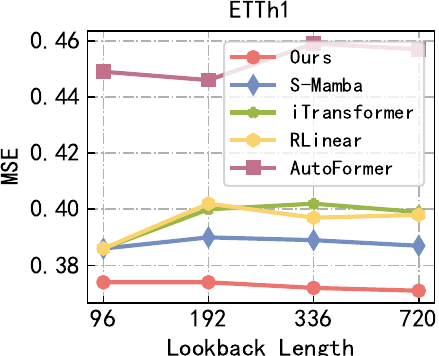}
  \end{minipage}\hspace{-3.mm}
  &
  \begin{minipage}{0.21\textwidth}
	\includegraphics[width=\textwidth]{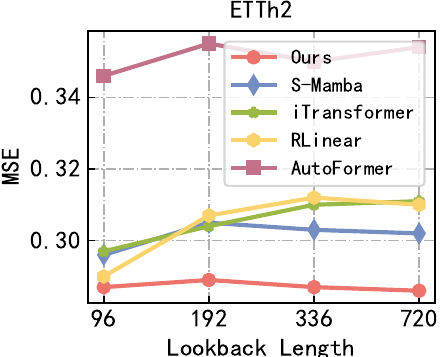}
  \end{minipage}\hspace{-3.0mm}
  \\\hspace{-4.0mm}
  \begin{minipage}{0.21\textwidth}
	\includegraphics[width=\textwidth]{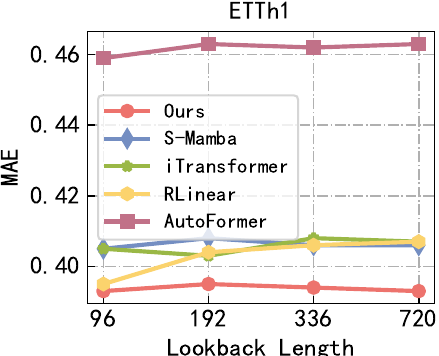}
  \end{minipage}\hspace{-3.mm}
  &
  \begin{minipage}{0.21\textwidth}
	\includegraphics[width=\textwidth]{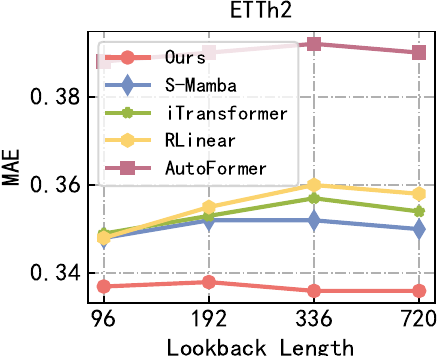}
  \end{minipage}\hspace{-3.0mm}\hspace{-3.0mm}
\end{tabular}
\caption{Long-term prediction with the lookback length from the range [96, 192, 336, 720]}
\label{fig:look_back}
\vspace*{-0.2in}
\end{figure}
\subsection{{\zqr{Long-term Prediction Comparison}} (Q4)}
This section investigates the effectiveness of our proposed framework, \model, in long-term time series prediction compared to other state-of-the-art methods. We conduct a comparative analysis against Transformer-based baselines (iTransformer, Rlinear, Autoformer) and a related Mamba-based method (S-Mamba). The results, presented in Figure~\ref{fig:look_back}, reveal the following key observations: \textbf{Superior Long-Term Performance of Mamba-Based Methods:} Compared to Transformer-based baselines, both S-Mamba and our method, \model, which are based on the Mamba architecture, demonstrate superior performance in terms of MAE and MSE. Furthermore, both methods exhibit a stable performance trend as the lookback window size increases from 96 to 720. This indicates that Mamba-based methods are more adept at capturing temporal patterns and dependencies, effectively preserving sequential features in long-term time series data. \textbf{Enhanced Long-Term Prediction with \model:}  Comparing \model to S-Mamba, our method shows a clear trend of reduced or maintained performance with increasing lookback window size. We attribute this improvement to the incorporation of the Fourier transform and the inverse Laplace transform, which effectively capture periodic dependencies and further enhance the ability to handle long-term prediction.
These findings highlight the effectiveness of \model in capturing complex temporal dynamics and maintaining performance even with extended lookback windows, demonstrating its significant advantage for long-term time series prediction.
\begin{figure}
\centering
\begin{tabular}{c c c}
\\\hspace{-4.0mm}
  \begin{minipage}{0.21\textwidth}
	\includegraphics[width=\textwidth]{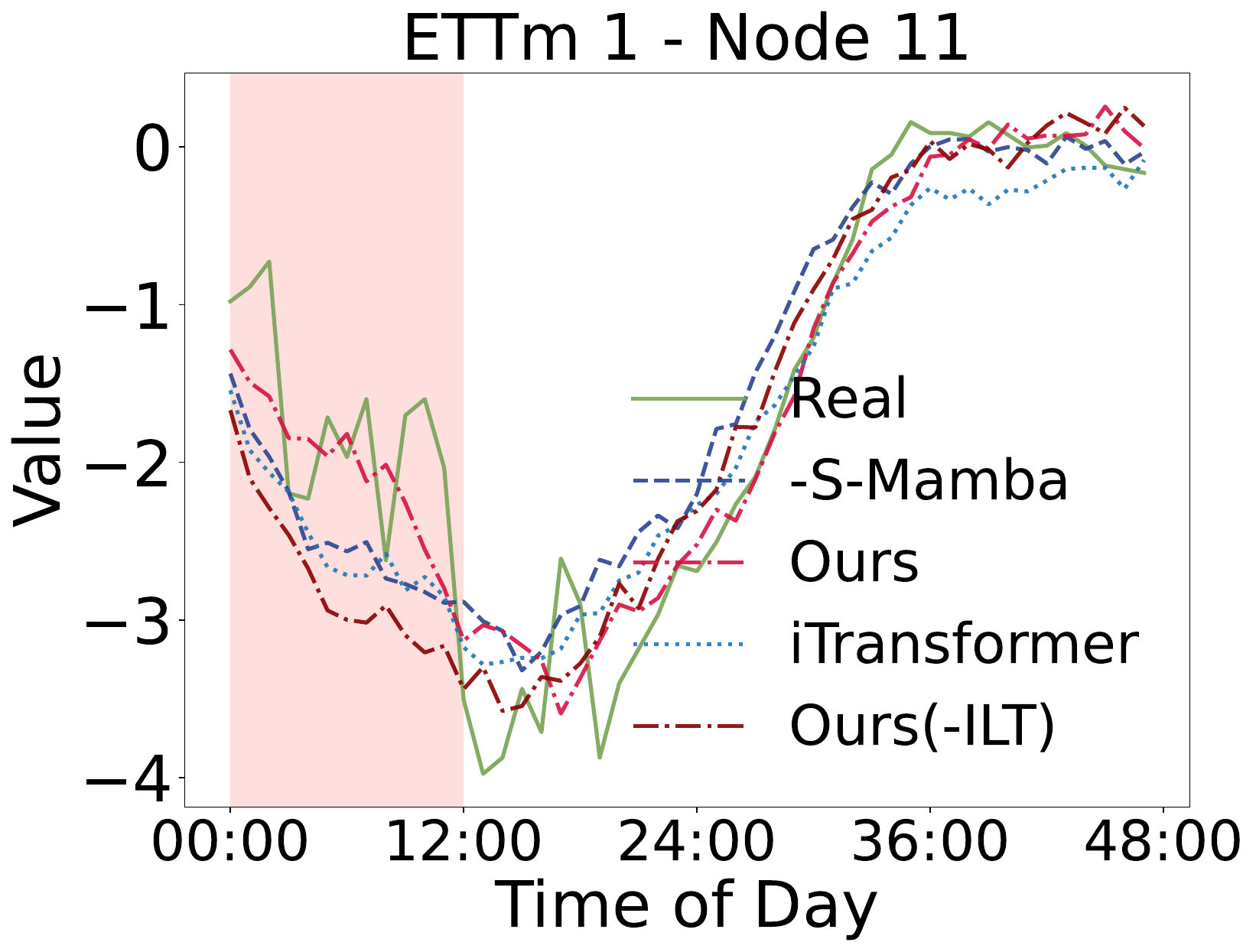}
  \end{minipage}\hspace{-3.mm}
  &
  \begin{minipage}{0.205\textwidth}
	\includegraphics[width=\textwidth]{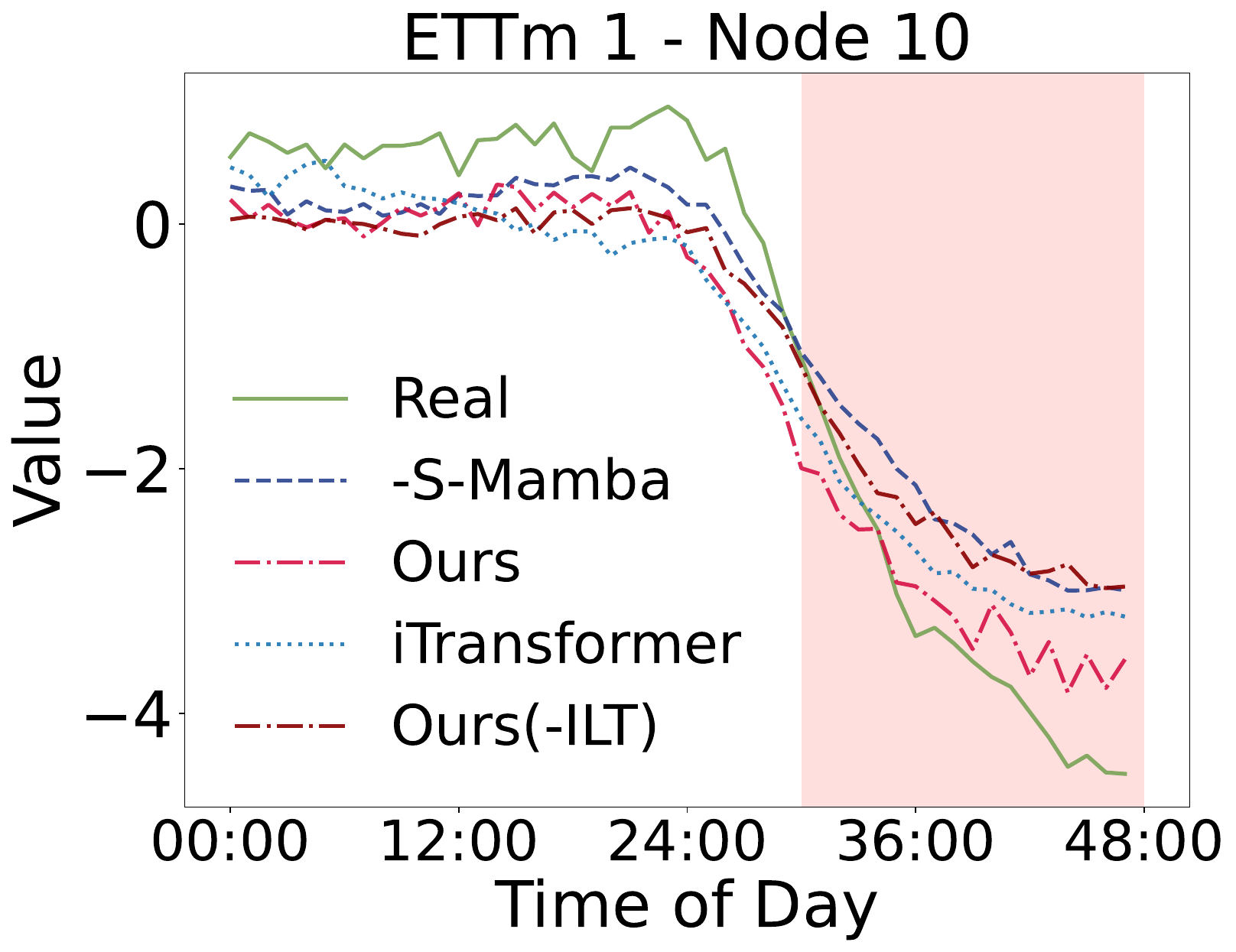}
  \end{minipage}\hspace{-3.0mm}
\end{tabular}
\caption{Case study of \model in terms of transient dynamics like short-term fluctuations.}
\vspace{-0.2in}
\label{fig:case_study2}
\end{figure}

\subsection{Case Study (Q5)}\label{sec:case_study}
\begin{figure*}
\vspace*{-4mm}
\centering
\begin{tabular}{c cc c}
\\\hspace{-4.0mm}
  \begin{minipage}{0.22\textwidth}
	\includegraphics[width=\textwidth]{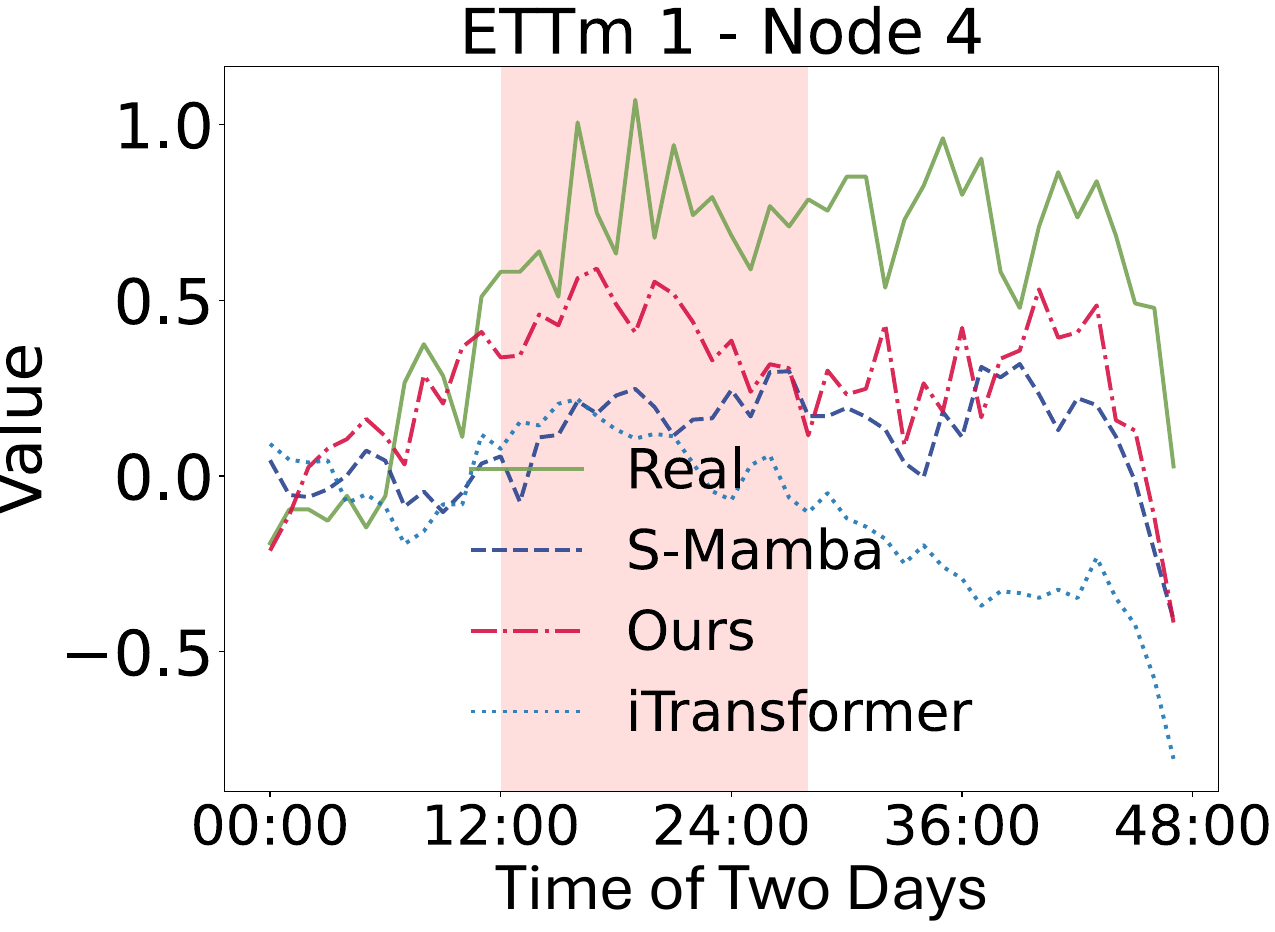}
  \end{minipage}\hspace{-3.mm}
  &
  \begin{minipage}{0.22\textwidth}
	\includegraphics[width=\textwidth]{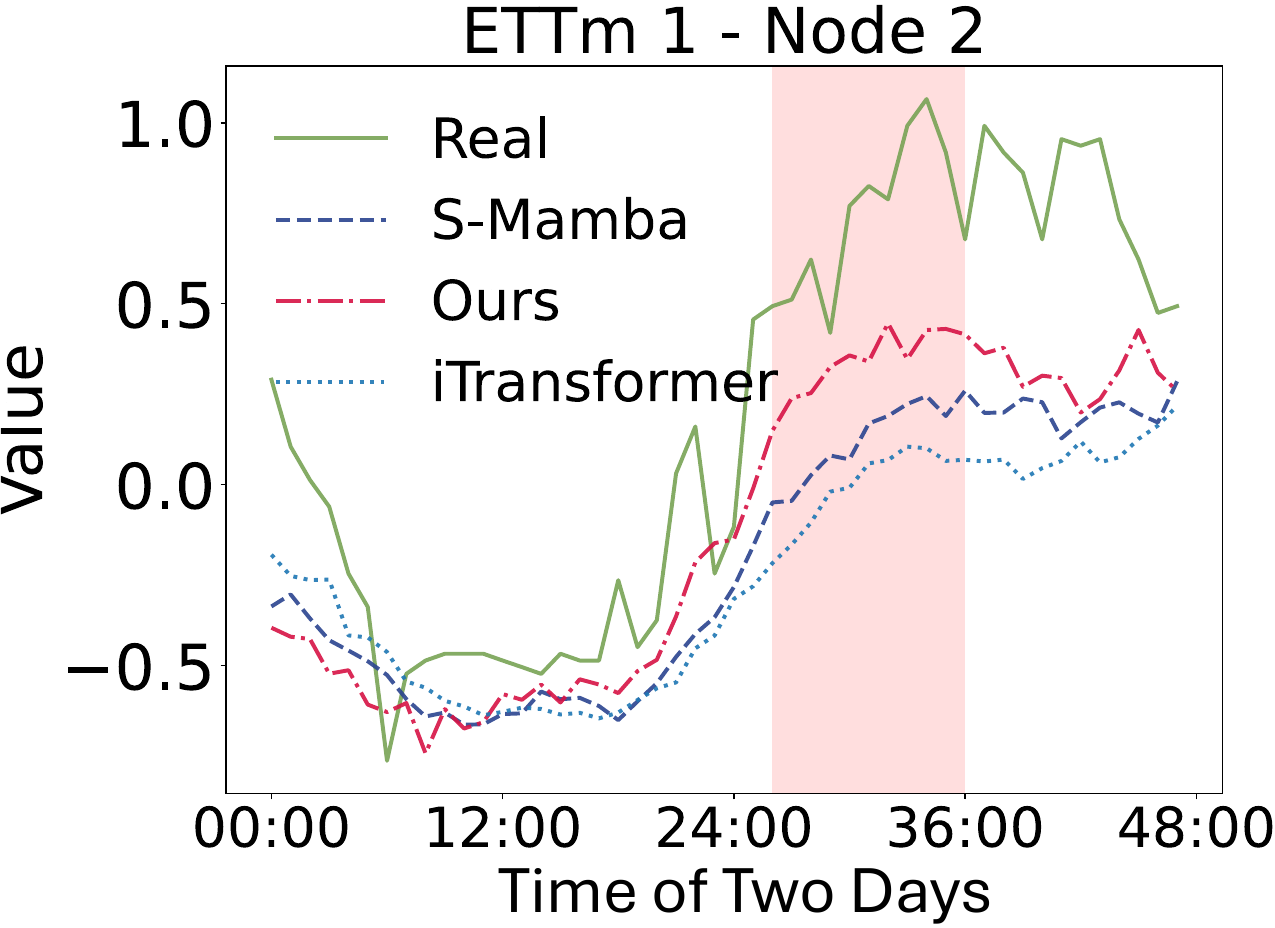}
  \end{minipage}\hspace{-3.0mm}
  &
  \begin{minipage}{0.22\textwidth}
	\includegraphics[width=\textwidth]{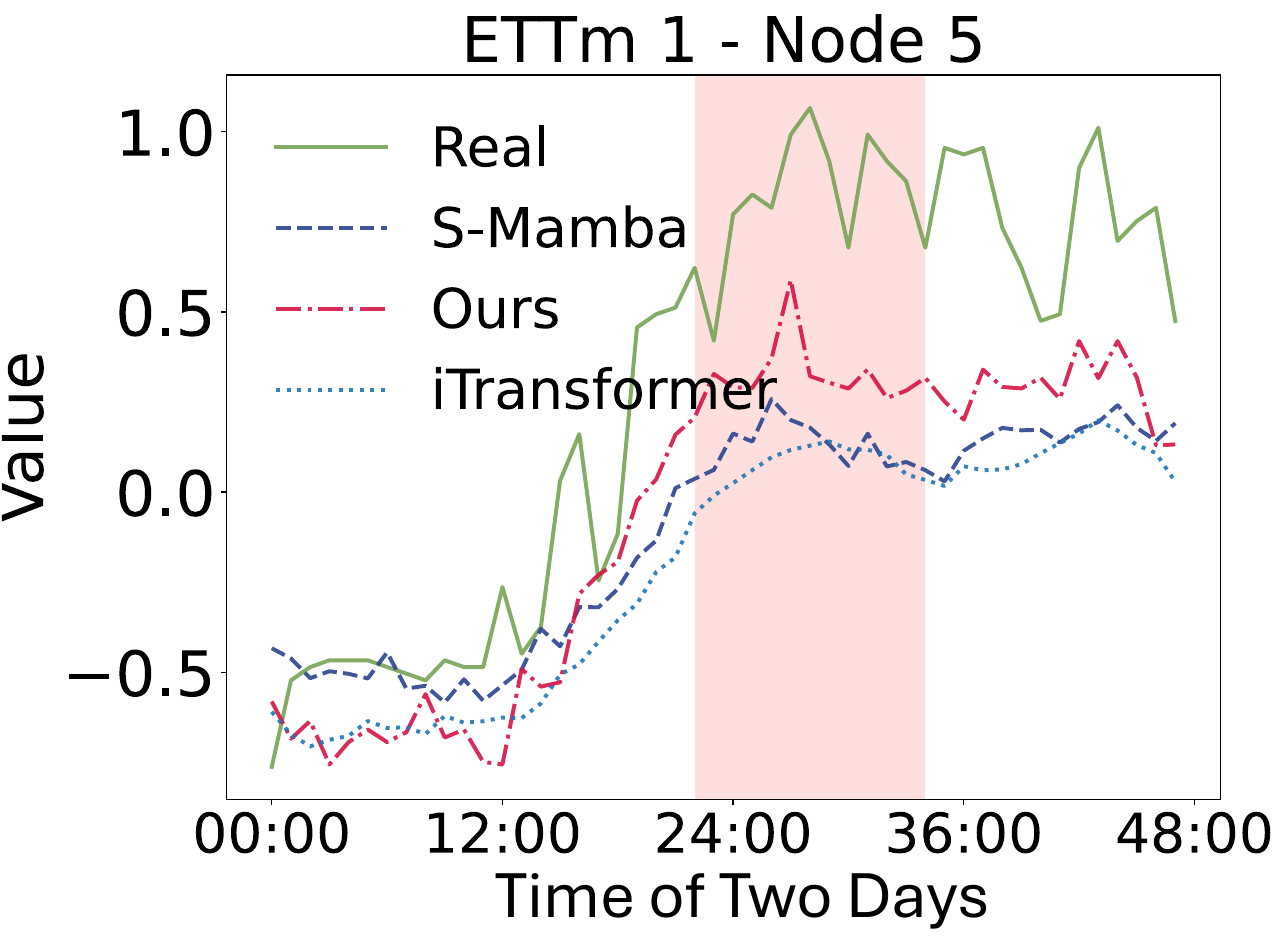}
  \end{minipage}\hspace{-3.0mm}
  &
  \begin{minipage}{0.22\textwidth}
	\includegraphics[width=\textwidth]{figures/pa4}
  \end{minipage}\hspace{-3.mm}
  \\\hspace{-4.0mm}
  \begin{minipage}{0.22\textwidth}
	\includegraphics[width=\textwidth]{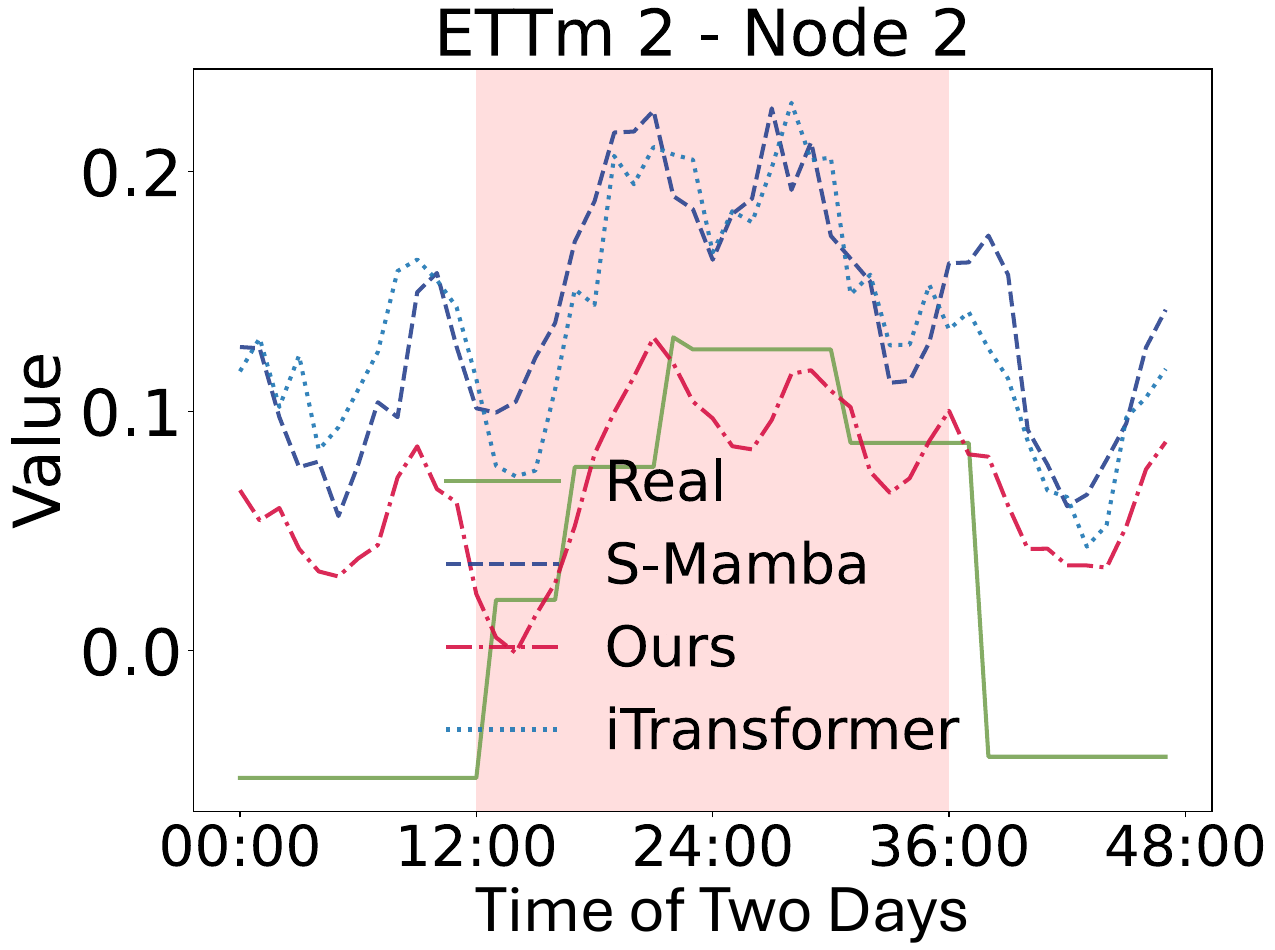}
  \end{minipage}\hspace{-3.0mm}
  &
  \begin{minipage}{0.22\textwidth}
	\includegraphics[width=\textwidth]{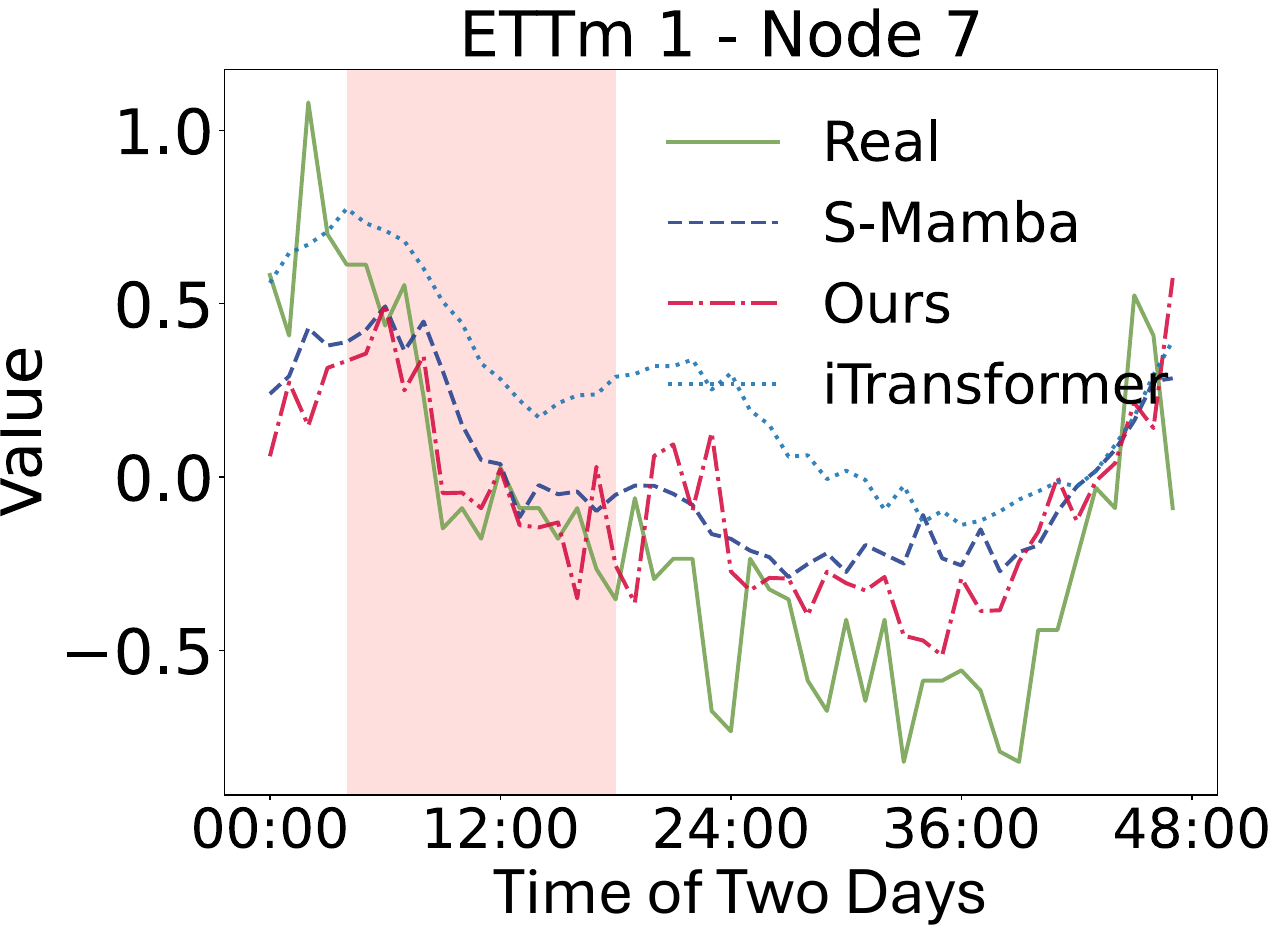}
  \end{minipage}\hspace{-3.0mm}
  &
  \begin{minipage}{0.22\textwidth}
	\includegraphics[width=\textwidth]{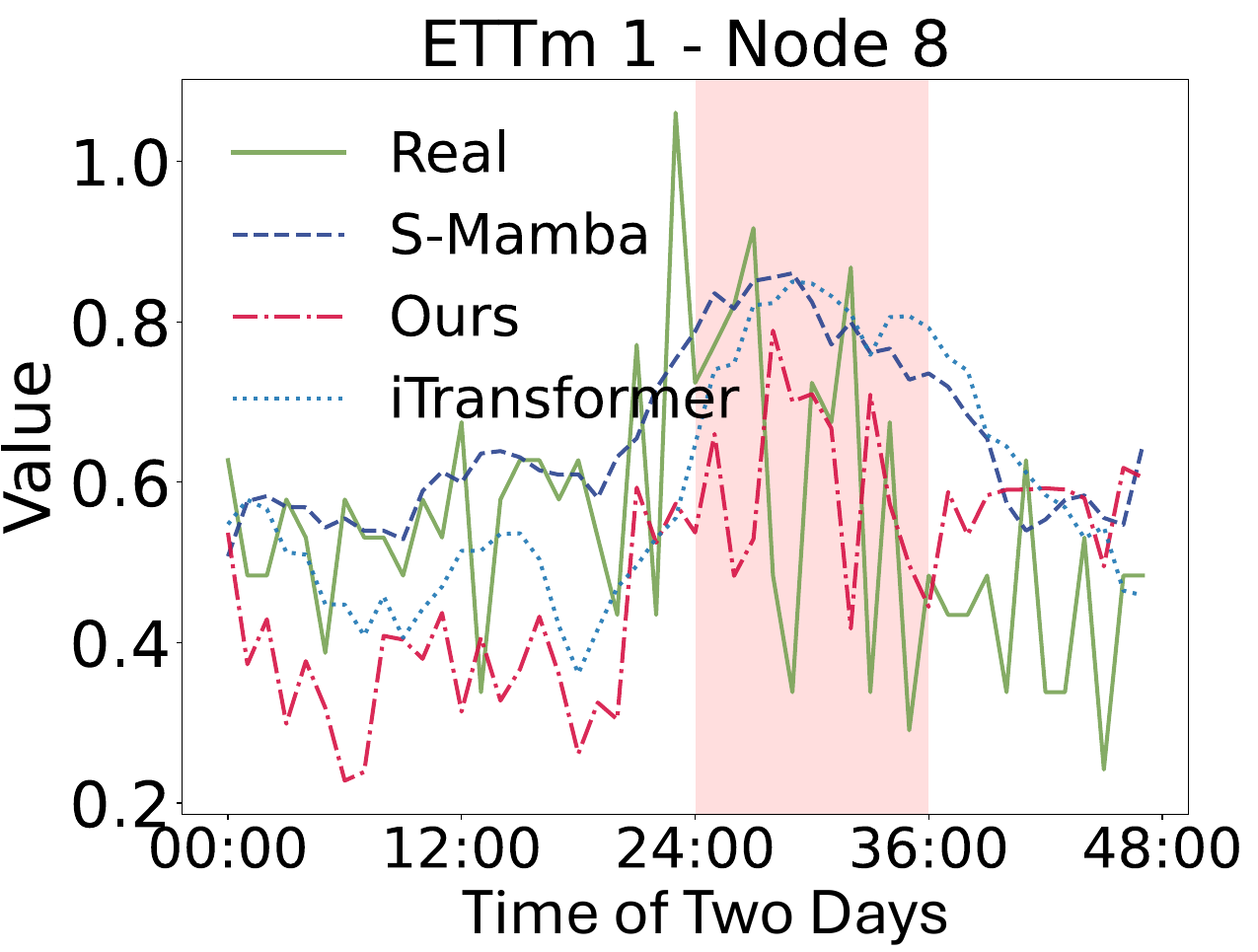}
  \end{minipage}\hspace{-3.mm}
  &
  \begin{minipage}{0.22\textwidth}
	\includegraphics[width=\textwidth]{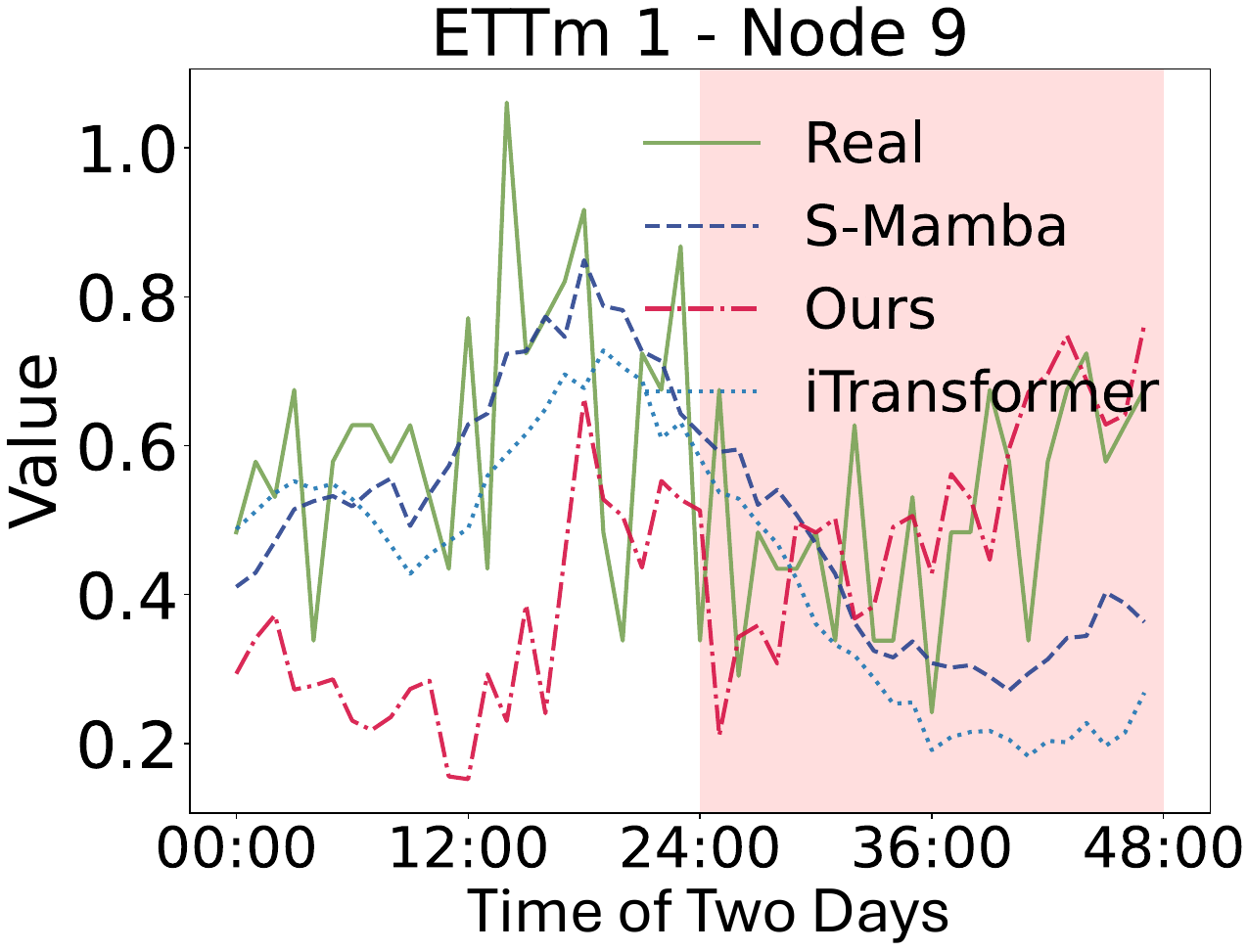}
  \end{minipage}\hspace{-3.0mm}

\end{tabular}
\caption{{\ICLRRevision{Case study of \model in terms of multi-scale periodicity.}}}
\vspace*{-4mm}
\label{fig:case_study1}
\end{figure*}

This section examines the efficacy of our proposed framework, \model, in capturing multi-scale periodicity and transient dynamics, particularly short-term fluctuations, within time series data. To illustrate its capabilities, we present case studies based on the ETTm1 and ETTm2 datasets, as depicted in Figure~\ref{fig:case_study2} and Figure~\ref{fig:case_study1}. These figures showcase the variations in the datasets over two consecutive days and 12 hours, respectively. For comparative analysis, we include the predicted results of two state-of-the-art baselines, S-Mamba and iTransformer. Each plot displays four curves: the ground truth values, the predictions generated by S-Mamba, the predictions from iTransformer, and the predictions obtained using our \model. We have the following observations: Figure~\ref{fig:case_study2} showcases the effectiveness of our method, \model, in capturing transient dynamics, particularly short-term fluctuations, compared to S-Mamba and iTransformer. The Laplace Transform within our framework significantly enhances its ability to model these dynamics, leading to improved performance. While S-Mamba demonstrates some capability in capturing transient dynamics, our method exhibits a more pronounced advantage. Conversely, iTransformer shows limited effectiveness in capturing these short-term fluctuations. {\tkde{We further evaluate our model by removing the inverse Laplace transform (ILT), as shown in Figure~\ref{fig:case_study2}. The results demonstrate degraded performance in predicting short fluctuations, confirming ILT's critical role in handling dynamics.}}

Figure~\ref{fig:case_study1} demonstrates \model’s superior ability to capture multi-scale periodicity in time-series data compared to S-Mamba and iTransformer, significantly improving prediction accuracy. The integrated Fourier and Laplace transforms enable nuanced modeling of periodic patterns, outperforming S-Mamba’s partial capability and iTransformer’s limited effectiveness. This validates our’s strength in decoding complex temporal structures, where traditional methods fall short.


\subsection{Efficiency (Q6)}
\begin{figure*}
\vspace*{-4mm}
\centering
\begin{tabular}{c c c c}
\\\hspace{-4.0mm}
  \begin{minipage}{0.22\textwidth}
	\includegraphics[width=\textwidth]{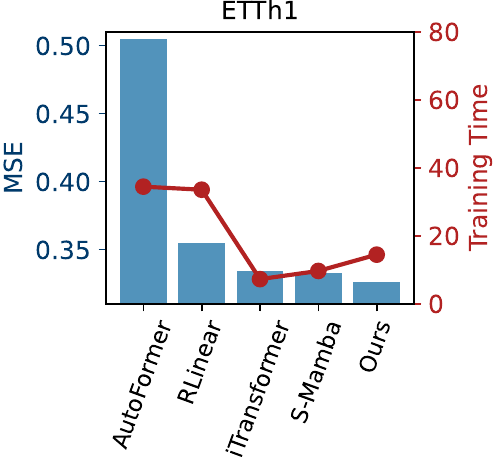}
  \end{minipage}\hspace{-3.mm}
  &
  \begin{minipage}{0.22\textwidth}
	\includegraphics[width=\textwidth]{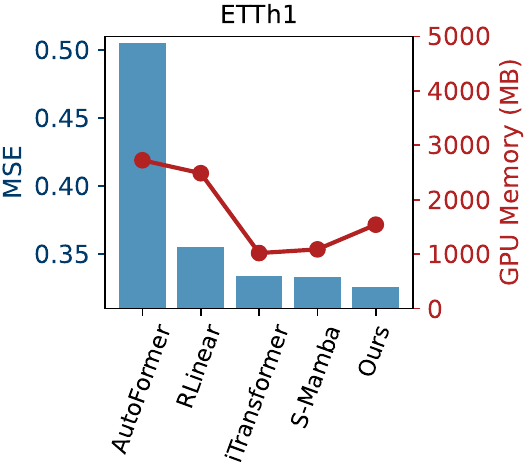}
  \end{minipage}\hspace{-3.0mm}
  &
  \begin{minipage}{0.22\textwidth}
	\includegraphics[width=\textwidth]{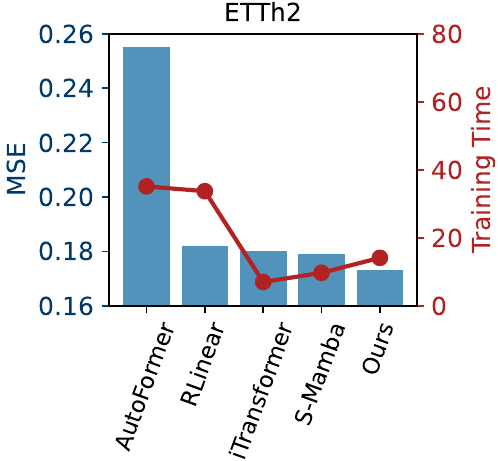}
  \end{minipage}\hspace{-3.0mm}
  &
  \begin{minipage}{0.22\textwidth}
	\includegraphics[width=\textwidth]{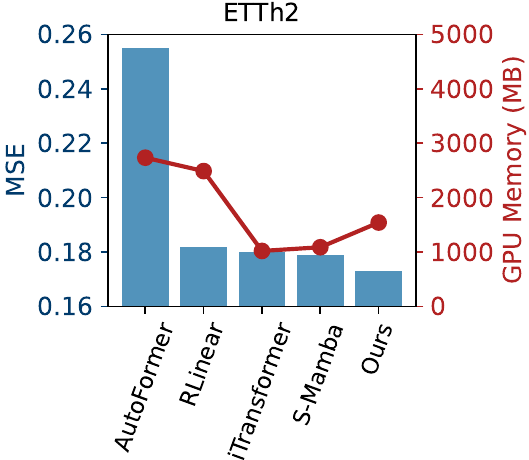}
  \end{minipage}\hspace{-3.0mm}
\end{tabular}
\vspace*{-1mm}
\caption{Model efficiency comparison on ETTh1 and ETTh2. The batch size is 32.}
\label{fig:efficiency2}
\vspace*{-2mm}
\end{figure*}

This section evaluates the computational efficiency of our proposed framework, \model, in comparison to several state-of-the-art baselines, including AutoFormer, RLinear, iTransformer, and S-Mamba. We assess efficiency on the ETTh1 and ETTh2 datasets, considering both training time per epoch and GPU memory consumption. The results, presented in Figure~\ref{fig:efficiency2}, demonstrate the following: \textbf{Comparative Efficiency of \model:} Our method, \model, exhibits a favorable balance between performance and computational efficiency, achieving comparable training times and GPU memory costs to baselines. \textbf{Efficiency of Mamba-Based Methods:} Mamba-based methods, including \model and S-Mamba, demonstrate a compelling advantage in terms of training time and GPU memory consumption compared to Transformer-based baselines such as AutoFormer. This suggests that Mamba-based architectures offer a more efficient approach for handling time series data.
These findings highlight the computational efficiency of our proposed framework, \model, while also emphasizing the potential benefits of Mamba-based architectures for addressing computational resource constraints in time series modeling.

\begin{figure*}
\centering
\begin{tabular}{c c c}
\\\hspace{-4.0mm}
  \begin{minipage}{0.25\textwidth}
	\includegraphics[width=\textwidth]{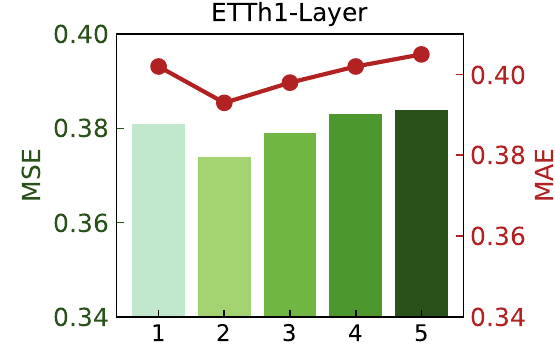}
  \end{minipage}\hspace{-3.mm}
  &
  \begin{minipage}{0.25\textwidth}
	\includegraphics[width=\textwidth]{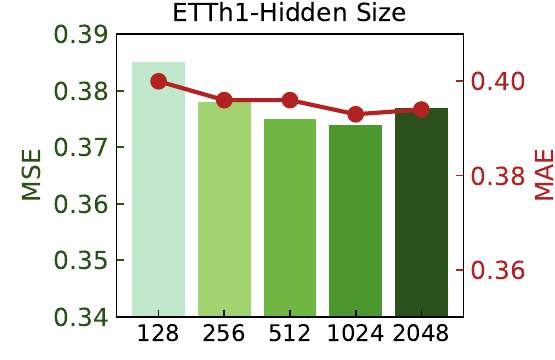}
  \end{minipage}\hspace{-3.0mm}
  &
  \begin{minipage}{0.25\textwidth}
	\includegraphics[width=\textwidth]{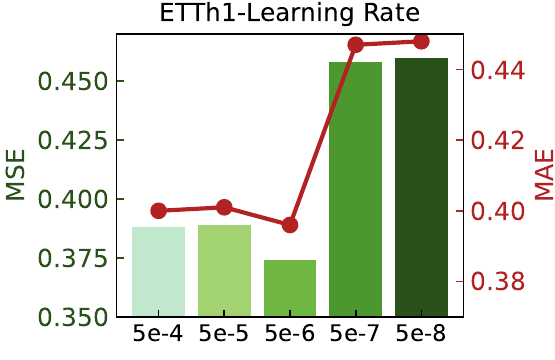}
  \end{minipage}\hspace{-3.0mm}
\end{tabular}
\caption{Hyperparameter study of \model.}
\vspace*{-2mm}
\label{fig:hyper}
\vspace*{-1mm}
\end{figure*}

\subsection{Hyperparameter Study (Q7)}
In this section, we aim to conduct a parameter study to evaluate the impact of important parameters on the performance of our model, \model. The results are presented in Figure~\ref{fig:hyper}. Specifically, we vary the number of \model\ layers within the range of $\left\{1,2,3,4,5\right\}$, the hidden size from $\left\{128,256,512,1024,2048\right\}$, and the learning rate from $\left\{5\times10^{-4},5\times10^{-5},5\times10^{-6},5\times10^{-7},5\times10^{-8}\right\}$. Based on the results, we provide a summary of observations regarding these three parameters and their effects on performance, measured by MSE and MAE metrics, as follows: \textbf{(1)} We examine the impact of \model\ layers on the performance of \model. We observe that \model\ achieves the best performance when the number of layers is set to 2. However, as we increase the number of \model\ layers, the performance starts to diminish. This suggests that additional layers may introduce an over-smoothing effect, which negatively affects the performance of \model. \textbf{(2)} We also conducted experiments to investigate the effect of hidden sizes on \model\ performance. We find that our model \model\ achieves the highest performance when the hidden size is set to 1024. This indicates that smaller hidden sizes may not provide sufficient information, while larger hidden sizes may introduce redundant information that hampers the performance of \model. \textbf{(3)} Furthermore, we examine the impact of the learning rate on performance and observe that our method \model\ achieves the best performance when the learning rate is set to $5\times10^{-6}$. Smaller or larger learning rates may result in insufficient convergence or overfitting, which adversely affects the performance.

\subsection{Long Lookback Comparison (Q8)}
\label{app:lookback_exp}
To evaluate the performance of various models on long lookback, we conducted experiments using a lookback length of 1500 on the ETTh1 and ETTh2 datasets. Table~\ref{tab:long_lookback} shows the MSE and MAE metrics for our proposed FLDmamba method, as well as other baseline models like S-Mamba, iTransformer, Rlinear, and AutoFormer. The results demonstrate that our FLDmamba method outperforms the other baselines across both datasets, highlighting its superior predictive capabilities.

\begin{table}[!ht]
    \centering
    \caption{\ICLRRevision{Performance of comparison when lookback length is set as 1500.}}\label{tab:long_lookback}
    \resizebox{0.35\textwidth}{!}{ 
    \begin{tabular}{c|cc|cc}
    \toprule
        & \multicolumn{2}{c|}{\ICLRRevision{ETTh1}} & \multicolumn{2}{c}{\ICLRRevision{ETTh2}}  \\ \midrule
        ~ & \ICLRRevision{MSE} & \ICLRRevision{MAE} & \ICLRRevision{MSE} & \ICLRRevision{MAE}  \\ \midrule
        \ICLRRevision{\model\textbf{(ours)}} & \ICLRRevision{\textbf{0.664}} & \ICLRRevision{\textbf{0.570}} & \ICLRRevision{\textbf{0.517}} & \ICLRRevision{\textbf{0.504}}  \\ \midrule
        \ICLRRevision{S-Mamba} & \ICLRRevision{0.715} & \ICLRRevision{0.603} & \ICLRRevision{0.539} & \ICLRRevision{0.522}  \\ \midrule
        \ICLRRevision{iTransformer} & \ICLRRevision{0.787} & \ICLRRevision{0.634} & \ICLRRevision{0.549} & \ICLRRevision{0.528}  \\ \midrule
        \ICLRRevision{Rlinear} & \ICLRRevision{1.281} & \ICLRRevision{0.884} & \ICLRRevision{3.015} & \ICLRRevision{1.366}  \\ \midrule
        \ICLRRevision{AutoFormer} & \ICLRRevision{0.687} & \ICLRRevision{0.614} & \ICLRRevision{0.648} & \ICLRRevision{0.575}  \\ \bottomrule
    \end{tabular}}
    \vspace*{-5mm}
\end{table}

\section{Conclusion}
\label{sec:conclusoin}
In conclusion, this paper addresses the limitations of existing time series prediction models, particularly in capturing multi-scale periodicity, transient dynamics and noise alleviation within long-term predictions. We propose a novel framework, \model, which leverages the strengths of both Fourier and Laplace transforms to effectively address these challenges. By integrating Fourier analysis into Mamba, \model enhances its ability to capture global-scale properties, such as multi-scale patterns, in the frequency domain. Our extensive experiments demonstrate that \model achieves state-of-the-art performance in most of cases on 9 datasets on time series prediction benchmarks. This work offers an effective and robust solution for long-term time series prediction, paving the way for its application in real-world scenarios. Future investigations will further enhance the model's adaptability to dynamic data environments.

\bibliographystyle{IEEEtran}
\bibliography{main}

\end{document}